\documentclass[pdflatex,sn-mathphys-num]{sn-jnl}

\usepackage{graphicx}%
\usepackage{multirow}%
\usepackage{amsmath,amssymb,amsfonts}%
\usepackage{amsthm}%
\usepackage{mathrsfs}%
\usepackage[title]{appendix}%
\usepackage{xcolor}%
\usepackage{textcomp}%
\usepackage{manyfoot}%
\usepackage{booktabs}%
\usepackage{algorithm}%
\usepackage{algorithmicx}%
\usepackage{algpseudocode}%
\usepackage{listings}%
\usepackage{float}
\usepackage{tabularx}
\usepackage{array}

\theoremstyle{thmstyleone}%
\theoremstyle{thmstyletwo}%
\theoremstyle{thmstylethree}%
\usepackage{placeins}

\newcolumntype{Y}{>{\raggedright\arraybackslash}X}
\newcolumntype{C}{>{\centering\arraybackslash}p{1.45cm}}

\begin{document}

\title[Dynamic Manipulation Hypergraphs for HAR]{Beyond Pairwise Relations: Dynamic Manipulation Hypergraphs for Vision-Based Human Activity Recognition}


\author*[1]{\fnm{Fatemeh} \sur{Ziaeetabar}}
\email{fziaeetabar@ut.ac.ir}

\affil[1]{\orgdiv{Department of Computer Science}, \orgname{School of Mathematics, Statistics and Computer Science, College of Science, University of Tehran}, \city{Tehran}, \state{Tehran}, \country{Iran}}

\abstract{Fine-grained recognition of manipulation activities requires explicit modeling of the evolving relations among hands, objects, tools, and supporting surfaces. Conventional graph-based methods represent these dependencies through pairwise edges, which may decompose a coordinated manipulation event into disconnected binary relations. We propose a dynamic manipulation hypergraph framework that represents multi-entity interaction configurations directly as higher-order relational units. At each temporal step, manipulation-relevant entities are encoded using appearance, spatial, motion, and semantic-role features. Manipulation-aware hyperedge candidates are then instantiated and ranked using proximity, contact, and motion-coupling predicates. A hypergraph reasoning network performs node-to-hyperedge and hyperedge-to-node message passing, followed by temporal attention over the evolving interaction structure. The framework additionally provides class-agnostic hyperedge-importance scores that identify the entity configurations and temporal intervals emphasized by the model, without interpreting them as causal explanations. Quantitative evaluation is conducted on EPIC-KITCHENS-100/VISOR and Assembly101 under an annotation-assisted entity-localization protocol. Video-only and entity-based methods provide contextual comparisons, whereas a matched pairwise graph and a static hypergraph constitute the principal controlled baselines because they use identical entity inputs and comparable relational settings. The proposed method improves HO-F1 over the matched pairwise graph by 6.9 percentage points on EPIC-KITCHENS-100/VISOR and 9.5 points on Assembly101, and exceeds the static hypergraph by 4.4 and 5.8 points, respectively. Complementary qualitative analysis on ARCTIC further illustrates the correspondence between highly ranked hyperedges and contact-rich manipulation intervals. These results demonstrate the value of time-varying higher-order relational modeling for fine-grained manipulation activity recognition.}


\maketitle
\section{Introduction}
\label{sec:introduction}

Vision-based human activity recognition in manipulation-heavy video is often determined not by clip-level appearance or motion alone, but by who interacts with what, how, and when. In egocentric kitchens, assembly, and contact-rich bimanual scenes, actions such as cutting, placing, fastening, or opening differ through coordinated relations among hands, manipulated objects, tools, and supporting surfaces. Although modern video foundation models learn strong global spatiotemporal representations \cite{Wang_2023_CVPR_VideoMAEv2,Wang_2024_ECCV_InternVideo2}, and graph-based methods model hand--object dependencies \cite{Kwon_2021_ICCV_H2O,Rodin_2024_CVPR_ActionSceneGraphs,Ziaeetabar_2024_IEEEAccess_HierarchicalBimanual,Ziaeetabar_2025_IEEEAccess_AdaptiveMultimodal}, most existing formulations still encode manipulation either implicitly in clip features or as collections of pairwise links. This can fragment a single manipulation event into local relations and obscure the joint configuration that determines the activity label.

Recent egocentric and manipulation-centered benchmarks have made this limitation increasingly visible. EPIC-KITCHENS-100 and VISOR pair long-form daily activities with fine-grained action labels, pixel-level masks, and hand--object relations \cite{Damen_2022_IJCV_EPIC100,Darkhalil_2022_NeurIPS_VISOR}; H2O, Assembly101, and ARCTIC further emphasize two-hand interaction, procedural activity, and dexterous contact-rich manipulation \cite{Kwon_2021_ICCV_H2O,Sener_2022_CVPR_Assembly101,Fan_2023_CVPR_ARCTIC}. Contact-centric and hand--object-centric approaches likewise show that explicit interaction cues matter for first-person understanding \cite{Dessalene_2021_arXiv_EgoOMG,Zhang_2022_arXiv_EgoHOS}. Hypergraphs provide a natural language for such higher-order structure because they can represent multi-entity configurations directly rather than decomposing them into binary edges \cite{Feng_2019_AAAI_HGNN}. Yet time-varying higher-order reasoning remains underexplored for fine-grained HAR.

We address this gap with a \emph{dynamic manipulation hypergraph} framework that models a video as a sequence of higher-order hyperedges among hands, manipulated objects, tools, and supporting surfaces. Candidate hyperedges are instantiated from manipulation templates,
ranked using proximity, contact, and motion-coupling predicates,
processed by the hypergraph reasoning network, and summarized through
temporal attention; the resulting class-agnostic hyperedge-importance
scores indicate the relational patterns emphasized by the model. We evaluate on EPIC-KITCHENS-100/VISOR and Assembly101 for quantitative recognition, and use ARCTIC for supporting qualitative contact analysis. The experimental protocol includes contextual comparisons with video-only baselines and controlled comparisons with a matched pairwise graph and a static hypergraph. Performance is
evaluated using accuracy, macro-F1, and higher-order action F1 (HO-F1), while controlled ablations examine the respective contributions of higher-order structure and time-varying hypergraph construction.

Our main contributions are:
\begin{itemize}
    \item We introduce a dynamic manipulation hypergraph representation for vision-based human activity recognition, in which time-varying higher-order hyperedges explicitly model coordinated interactions among hands, manipulated objects, tools, and supporting surfaces.

    \item We develop a manipulation-aware hypergraph construction strategy that integrates appearance, spatial, motion, and semantic-role features with proximity, contact, and motion-coupling predicates to rank candidate hyperedges and generate class-agnostic hyperedge-importance scores.

    \item We conduct controlled comparisons with a matched pairwise graph and a static hypergraph using identical entity inputs, temporal sampling, and comparable reasoning settings, thereby isolating the respective contributions of higher-order relational modeling and dynamic hypergraph construction. Video-only baselines are included as contextual comparisons.

    \item We validate the proposed framework on EPIC-KITCHENS-100/VISOR and Assembly101 using accuracy, macro-F1, and HO-F1. The method improves HO-F1 over the matched pairwise graph by 6.9 and 9.5 percentage points, respectively, while complementary qualitative analysis on ARCTIC illustrates the temporal alignment between highly ranked hyperedges and contact-rich manipulation intervals.
\end{itemize}

The current formulation still depends on reliable entity localization and a predefined template set. This suggests natural future directions in data-driven hyperedge discovery, richer object-state modeling, and more open-vocabulary manipulation reasoning.

\section{Related Work}
\label{sec:related_work}

Fine-grained manipulation activities are often difficult to recognize from appearance and motion alone, because their meaning depends on the evolving relations among hands, objects, tools, and surrounding surfaces. This section reviews the most relevant work on vision-based human activity recognition, graph and hypergraph learning for action modeling, and manipulation-centered human--object interaction understanding.

\subsection{Vision-Based Human Activity Recognition}
\label{subsec:vision_based_har}

Vision-based human activity recognition has recently benefited from large-scale video pretraining and Transformer-based architectures. Models such as VideoMAE~V2~\cite{Wang_2023_CVPR_VideoMAEv2} and InternVideo2~\cite{Wang_2024_ECCV_InternVideo2} have shown that strong video representations can be transferred to a wide range of recognition and video-language tasks. These models are effective in capturing global appearance and temporal patterns, but fine-grained manipulation activities often require more explicit reasoning about hands, objects, tools, and their evolving relations.

Egocentric video datasets have been especially important for this problem. EPIC-KITCHENS-100 provides large-scale first-person recordings of daily activities~\cite{Damen_2022_IJCV_EPIC100}, while VISOR adds pixel-level annotations of hands, active objects, and object relations~\cite{Darkhalil_2022_NeurIPS_VISOR}. More recent datasets such as HD-EPIC further increase the semantic and temporal detail of egocentric annotations~\cite{Perrett_2025_CVPR_HDEPIC}. These developments indicate that modern HAR is moving beyond coarse clip-level labels toward more structured descriptions of activity. However, many recognition models still represent the video mainly as a sequence of visual features, leaving the relational structure of manipulation only implicitly encoded.

\subsection{Graph and Hypergraph Learning for Action Recognition}
\label{subsec:graph_hypergraph_learning}

Graph neural networks provide a natural way to represent structured action data. In skeleton-based recognition, ST-GCN introduced a widely adopted formulation in which body joints are modeled as nodes and anatomical connections as spatial-temporal edges~\cite{Yan_2018_AAAI_STGCN}. Later graph-based models improved this idea using adaptive topologies and more expressive relational aggregation. Recent methods such as 3Mformer~\cite{Wang_2023_CVPR_3Mformer} and BlockGCN~\cite{Zhou_2024_CVPR_BlockGCN} show that explicit structural modeling remains useful even in the era of Transformer-based action recognition.

A limitation of ordinary graphs is that they encode relations as pairwise edges. This can be restrictive when an action is defined by a group of entities rather than by isolated pairwise relations. Hypergraph models address this issue by allowing one hyperedge to connect multiple nodes. Hyperformer~\cite{Zhou_2022_arXiv_Hyperformer} and Autoregressive Adaptive Hypergraph Transformer~\cite{Ray_2025_WACV_AutoregAdHGFormer} have explored this idea mainly for skeleton-based activity recognition. More recently, HyperGLM used hypergraph structures for video scene graph generation and anticipation~\cite{Nguyen_2025_CVPR_HyperGLM}. These works support the value of higher-order modeling, but most of them remain focused on skeleton joints or generic scene entities. In contrast, our work studies higher-order relations specifically in manipulation scenes, where meaningful units often involve hands, objects, tools, surfaces, and contact-state transitions.

\subsection{Manipulation and Human--Object Interaction Understanding}
\label{subsec:manipulation_hoi}

Manipulation understanding is closely related to human--object interaction recognition, but it requires more detailed temporal and relational modeling. H2O introduced a first-person setting for two-hand object manipulation~\cite{Kwon_2021_ICCV_H2O}. Assembly101 provides large-scale procedural videos of assembly and disassembly activities~\cite{Sener_2022_CVPR_Assembly101}, while HOI4D and ARCTIC offer richer egocentric, 3D, and bimanual hand--object interaction annotations~\cite{Liu_2022_CVPR_HOI4D,Fan_2023_CVPR_ARCTIC}. These datasets show that manipulation activities are increasingly studied as structured interactions rather than simple action labels.

Recent methods have also emphasized contact, object state, and temporal scene structure. Shiota et al.~\cite{Shiota_2024_WACV_ContactState} showed that hand--object contact and object-state changes are useful cues for egocentric action recognition. Action Scene Graphs represent long-form egocentric videos using evolving graphs of actions, objects, and relations~\cite{Rodin_2024_CVPR_ActionSceneGraphs}. These works are highly relevant to our motivation, but they mainly rely on pairwise graph structures or action-object relations.

Our earlier work progressively developed structured representations for manipulation understanding. Complex manipulation videos were first decomposed into semantically coherent multi-sentence descriptions~\cite{Ziaeetabar_2024_MVA_MultiSentence}. Hierarchical graph modeling was subsequently introduced for the recognition and description of bimanual actions~\cite{Ziaeetabar_2024_IEEEAccess_HierarchicalBimanual},
followed by adaptive multimodal graph reasoning incorporating
foundation-model features for fine-grained action recognition~\cite{Ziaeetabar_2025_IEEEAccess_AdaptiveMultimodal}.
Complementary work on spatial-semantic Graph-Transformer architectures and structural pruning further demonstrated the utility of structured graph representations in dense visual
reasoning~\cite{Ziaeetabar2025EfficientGFormer}. Building on these
graph-based developments, the present work advances from pairwise
relational modeling to dynamic manipulation hypergraphs, representing configurations such as \emph{hand--cup--table},
\emph{hand--knife--food--board}, and \emph{left hand--right hand--object} directly as higher-order interaction units.
\section{Proposed Method}
\label{sec:method}

The proposed method represents a manipulation video as a sequence of dynamic hypergraphs. 

Each hypergraph contains manipulation-relevant entities selected from five canonical roles:
left hand, right hand, manipulated object, tool, and supporting surface.
Its hyperedges encode higher-order interaction configurations among these entities.

This section first defines the recognition problem and the overall
pipeline, then describes visual entity representation, hypergraph
construction, hypergraph reasoning, and the final activity prediction
with class-agnostic hyperedge-importance scores.

\subsection{Overview and Problem Formulation}
\label{subsec:method_overview}

Given an input video clip \(X=\{x_t\}_{t=1}^{T}\), where \(x_t\) denotes the \(t\)-th frame or short temporal segment, the objective is to predict an activity label \(y \in \mathcal{Y}\). In manipulation-centered HAR, the label is often determined not only by the appearance or motion of individual entities, but also by how several entities jointly participate in an interaction. For example, \emph{placing} involves a hand, an object, and a supporting surface, while \emph{cutting} may involve a hand, a tool, a target object, and a board. This motivates a representation that can preserve the structure of such multi-entity configurations.

Our previous graph-based studies modeled manipulation videos through structured relational representations for bimanual action recognition, description generation, and fine-grained action understanding~\cite{Ziaeetabar_2024_IEEEAccess_HierarchicalBimanual,Ziaeetabar_2025_IEEEAccess_AdaptiveMultimodal}. Those formulations showed the benefit of explicit graph reasoning, but they still represented interactions mainly through pairwise edges. In the present work, we extend this idea by representing each video as a sequence of dynamic manipulation hypergraphs,
\begin{equation}
    \mathcal{G}(X)=\{\mathcal{H}_t\}_{t=1}^{T},
\end{equation}
where each hypergraph is defined as
\begin{equation}
    \mathcal{H}_t=(\mathcal{V}_t,\mathcal{E}_t).
\end{equation}
Here, \(\mathcal{V}_t=\{v_1^t,\ldots,v_{N_t}^t\}\) denotes the set of manipulation-relevant entities at time
\(t\), selected from five canonical entity roles: left hand, right hand, manipulated object, tool, and supporting surface. 
The set \(\mathcal{E}_t=\{e_1^t,\ldots,e_{M_t}^t\}\) contains hyperedges, where each hyperedge \(e_j^t \subseteq \mathcal{V}_t\) connects two or more entities that jointly form a meaningful manipulation configuration.

The recognition problem is therefore formulated as learning a mapping from a sequence of dynamic manipulation hypergraphs to a video-level representation:
\begin{equation}
    f_{\theta}: \{\mathcal{H}_t\}_{t=1}^{T} \rightarrow \mathbb{R}^{d},
\end{equation}
where \(f_{\theta}\) denotes the hypergraph reasoning model and \(d\) is the dimensionality of the learned representation. The output of this mapping is the final video-level feature vector \(\mathbf{z} \in \mathbb{R}^{d}\), which is subsequently used by the classification head described in Section~\ref{subsec:prediction_explanation}.

The model first performs higher-order reasoning within each temporal step and then aggregates the resulting representations over time. Unlike a standard graph, which decomposes an interaction into several pairwise relations, the proposed hypergraph can encode configurations such as \emph{hand--object--surface} or \emph{hand--tool--object--surface} as single relational units. This allows the model to represent manipulation patterns whose meaning depends on the joint participation of multiple entities, rather than on isolated pairwise connections alone.

\begin{figure}[t]
    \centering
    \includegraphics[width=\linewidth]{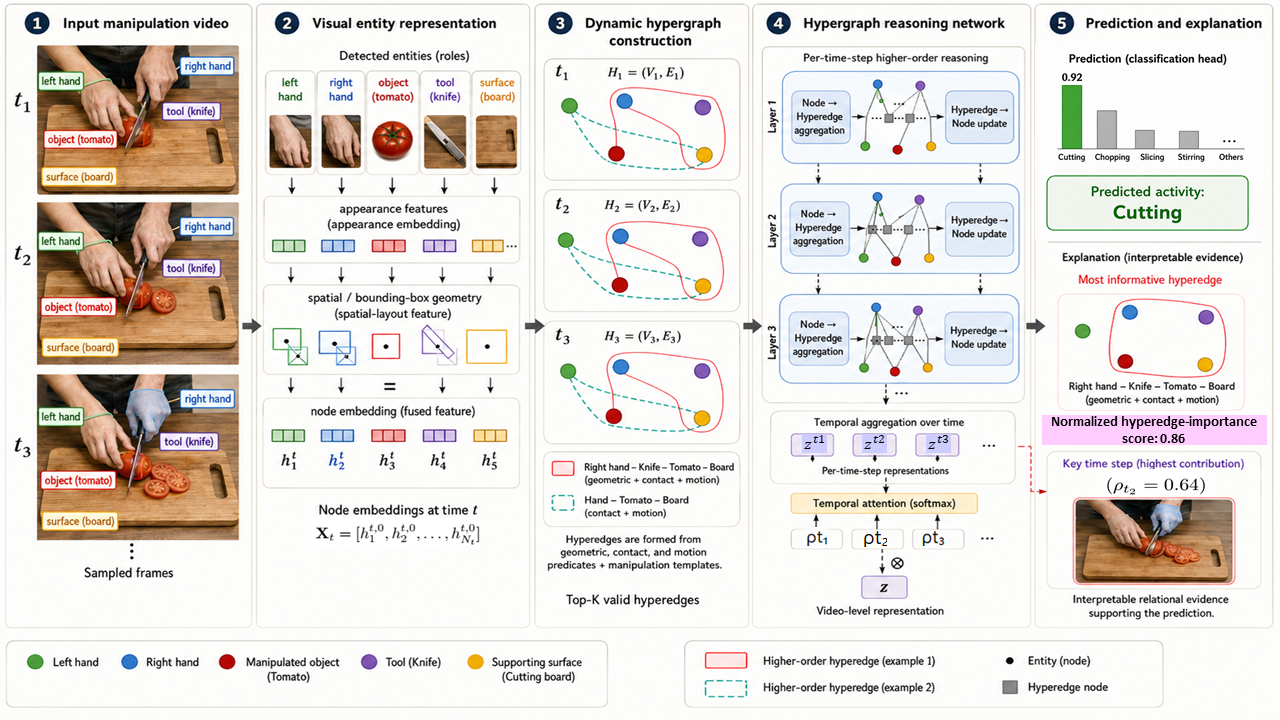}
    \caption{Overview of the proposed dynamic manipulation hypergraph
framework. Manipulation-relevant entities are organized into
time-varying hyperedges and processed by the hypergraph reasoning
network for activity prediction. The displayed hyperedge-importance
score is normalized within the clip for visualization.}
    \label{fig:method_overview}
\end{figure}

Figure~\ref{fig:method_overview} illustrates the overall pipeline of the proposed framework. The input video is first converted into manipulation-relevant entity representations. These entities are then used to construct a sequence of dynamic hypergraphs, which are processed by a hypergraph reasoning network. The resulting video-level representation is used for activity classification, while the top-ranked hyperedges indicate the relational patterns emphasized by the model.

\subsection{Entity Extraction, Role Assignment, and Visual Representation}
\label{subsec:entity_representation}

At each temporal step \(t\), the input frame is represented by a
variable-size set of manipulation-relevant entities drawn from five
canonical functional roles: left hand, right hand, manipulated object,
tool, and supporting surface. The available entities form the node set
\(\mathcal{V}_t=\{v_i^t\}_{i=1}^{N_t}\), where \(N_t\leq 5\). Because
some roles may be absent or temporarily occluded, the number of nodes
can vary across temporal steps.

For each sampled frame, a fixed entity-extraction pipeline produces a
set of candidate instance masks,
\begin{equation}
\mathcal{C}_t=
\left\{
(M_k^t,l_k^t,p_k^t)
\right\}_{k=1}^{K_t},
\end{equation}
where \(M_k^t\), \(l_k^t\), and \(p_k^t\) denote the mask, semantic
category, and confidence of candidate \(k\), respectively. Dataset
annotations are used when specified by the experimental protocol;
otherwise, the masks are generated using the pretrained detection and
instance-segmentation pipeline described in
Section~\ref{subsec:implementation_details}. The extraction models are
fixed and are not updated during activity-recognition training.

Entity identities are preserved across sampled frames using
dataset-provided track identifiers when available. Otherwise,
detections in consecutive frames are associated using mask
intersection-over-union, appearance similarity, and normalized
centre displacement. This temporal association prevents the motion
descriptor from being calculated between unrelated instances.

The candidate entities are assigned to functional roles independently
of the ground-truth or predicted activity label. Left and right hands
are determined from the corresponding annotation or detector labels.
Among the remaining object candidates, the manipulated object is
selected using label-independent interaction evidence, including
hand--object contact, spatial proximity, motion coupling, detection
confidence, and temporal persistence. Tool candidates are identified
using a predefined semantic category vocabulary established from the
training data before test evaluation. When multiple tool candidates
are present, the candidate exhibiting the strongest interaction with
a hand and the selected manipulated object is retained. Supporting
surfaces are selected from predefined planar-region categories using
object--surface proximity, relative vertical position, and spatial
overlap.

For two candidate instances indexed by \(i\) and \(j\) in
\(\mathcal{C}_t\), we define the interaction score
\begin{equation}
I_{ij}^t
=
\frac{1}{3}
\left(
q_{ij}^t+\pi_{ij}^t+\kappa_{ij}^t
\right),
\end{equation}
where the contact, proximity, and motion-coupling scores are defined in
Section~\ref{subsec:hypergraph_construction}. Let
\(\mathcal{C}_t^{\mathrm{hand}}\) denote the candidate instances
identified as hands.

For candidate object track \(o\) observed at temporal step \(t\), define
\begin{equation}
J_o^t
=
\begin{cases}
\displaystyle
\max_{h\in\mathcal{C}_t^{\mathrm{hand}}} I_{ho}^t,
&
\mathcal{C}_t^{\mathrm{hand}}\neq\varnothing,\\[2mm]
0,
&
\mathcal{C}_t^{\mathrm{hand}}=\varnothing.
\end{cases}
\end{equation}
The frame-level object-role score is
\begin{equation}
S_{\mathrm{obj}}(o,t)
=
\frac{1}{3}
\left(
p_o^t+J_o^t+u_o
\right),
\label{eq:object_role_score}
\end{equation}
where \(p_o^t\) is the candidate confidence and
\[
u_o
=
\frac{|\mathcal{T}_o|}{T},
\qquad
\mathcal{T}_o
=
\{t\in\{1,\ldots,T\}\mid o\text{ is observed at }t\},
\]
is the temporal-persistence score of track \(o\). For
annotation-provided masks without a detector confidence, \(p_o^t\) is
set to \(1\).

The temporally averaged object-role score is
\begin{equation}
\bar{S}_{\mathrm{obj}}(o)
=
\frac{1}{|\mathcal{T}_o|}
\sum_{t\in\mathcal{T}_o}
S_{\mathrm{obj}}(o,t),
\end{equation}
and the manipulated-object track is selected as
\begin{equation}
o^*
=
\arg\max_o
\bar{S}_{\mathrm{obj}}(o).
\end{equation}
When several tracks are eligible for the same functional role, the
track with the highest temporally averaged label-independent
role-assignment score is retained. At temporal steps where the selected
track is not observed, the corresponding node is omitted.

When multiple candidates satisfy the same functional role, the
highest-scoring temporally persistent candidate is retained. If a
required entity cannot be detected at a particular temporal step, the
corresponding node is omitted and hyperedge templates requiring that
role are not instantiated. The video clip itself is not removed from
training or evaluation. Therefore, failures of the entity-extraction
pipeline remain reflected in the reported recognition results. Most
importantly, ground-truth activity labels, predicted class
probabilities, and test-set class statistics are never used for
entity-role assignment.

Each selected mask is converted into a tight bounding box and used to
crop the corresponding image region. The crop is passed through a
pretrained visual backbone to obtain an appearance descriptor
\(\mathbf{a}_i^t\). The spatial descriptor is calculated from the
normalized bounding-box coordinates, centre position, and mask area:
\begin{equation}
\mathbf{b}_{i}^{t}
=
\left[
\frac{x_{1,i}^{t}}{W},\;
\frac{y_{1,i}^{t}}{H},\;
\frac{x_{2,i}^{t}}{W},\;
\frac{y_{2,i}^{t}}{H},\;
\frac{c_{x,i}^{t}}{W},\;
\frac{c_{y,i}^{t}}{H},\;
\frac{A_i^t}{WH}
\right],
\end{equation}
where \(W\) and \(H\) are the frame width and height,
\((c_{x,i}^{t},c_{y,i}^{t})\) is the centre of entity \(i\), and
\(A_i^t\) is its mask area.

For entities associated across consecutive temporal steps, the motion
descriptor is calculated as the normalized displacement of the entity
centre:
\begin{equation}
\mathbf{m}_{i}^{t}
=
\left[
\frac{c_{x,i}^{t}-c_{x,i}^{t-1}}{W},\;
\frac{c_{y,i}^{t}-c_{y,i}^{t-1}}{H}
\right].
\end{equation}
Let \(r_i^t\in\{0,1\}\) indicate whether entity \(i\) has a valid
correspondence in the preceding sampled frame. The motion vector is set
to zero when \(r_i^t=0\), including at the first temporal step. This
zero vector is used only as a placeholder for unavailable motion.

A learnable semantic embedding \(\mathbf{s}_i^t\) represents the
assigned functional role rather than the ground-truth activity class.
The initial node representation is obtained by concatenating the
appearance, spatial, motion, and role descriptors:
\begin{equation}
\mathbf{h}_{i}^{t,0}
=
\phi_{\mathrm{ent}}
\left(
\left[
\mathbf{a}_{i}^{t};
\mathbf{b}_{i}^{t};
\mathbf{m}_{i}^{t};
\mathbf{s}_{i}^{t}
\right]
\right),
\end{equation}
where \(\phi_{\mathrm{ent}}\) is a multilayer perceptron and
\([\cdot;\cdot]\) denotes feature concatenation.

For example, a frame may contain a right hand, knife, tomato, and
cutting board. Based exclusively on their semantic categories and
observed interaction evidence, these entities may respectively be
assigned the roles of right hand, tool, manipulated object, and
supporting surface.

The right-hand node may subsequently be selected as the active hand
\(h_t^*\) for the tool-mediated template. Their representations encode what entities are
present, where they are located, how they move, and which functional
roles they currently fulfil before the manipulation hyperedges are
constructed.

\subsection{Dynamic Manipulation Hypergraph Construction}
\label{subsec:hypergraph_construction}

After obtaining the node representations, we construct a dynamic
manipulation hypergraph at each temporal step. The purpose of this
stage is to convert isolated entity nodes into higher-order interaction
units that are meaningful for manipulation-centered HAR. Instead of
connecting entities only through pairwise edges, we form hyperedges
that represent complete manipulation configurations, such as a hand
acting on an object over a surface, or a hand using a tool on an object.

For each temporal step \(t\), we first compute geometric, contact, and
motion-based predicates between entity pairs. Let
\(c_i^t=(c_{x,i}^t,c_{y,i}^t)\) denote the center of entity \(v_i^t\),
obtained from its mask or bounding box. The normalized center distance
between two entities is defined as
\begin{equation}
d_{ij}^{t}
=
\frac{\lVert c_i^t-c_j^t\rVert_2}
{\sqrt{W^2+H^2}},
\end{equation}
where \(W\) and \(H\) are the frame width and height. The corresponding
proximity score is defined as
\begin{equation}
\pi_{ij}^{t}
=
\exp\left(-\frac{d_{ij}^{t}}{\tau_d}\right),
\end{equation}
where \(\tau_d>0\) is a proximity scale controlling sensitivity to
center distance. Consequently, \(\pi_{ij}^{t}\in(0,1]\), with larger
values indicating greater spatial proximity.

Contact evidence is computed from the minimum distance between the two
entity masks. Let \(\mathcal{P}(M_i^t)\) denote the set of pixel
coordinates belonging to mask \(M_i^t\). The normalized mask distance
is defined as
\begin{equation}
\delta_{ij}^{t}
=
\frac{
\displaystyle
\min_{\mathbf{p}\in\mathcal{P}(M_i^t),
      \mathbf{q}\in\mathcal{P}(M_j^t)}
\lVert\mathbf{p}-\mathbf{q}\rVert_2
}{
\sqrt{W^2+H^2}
}.
\end{equation}
The continuous contact score is then calculated as
\begin{equation}
q_{ij}^{t}
=
\exp\left(-\frac{\delta_{ij}^{t}}{\tau_c}\right),
\end{equation}
where \(\tau_c>0\) is the contact-distance scale. Thus,
\(q_{ij}^{t}\in(0,1]\). Overlapping or directly touching masks have
\(\delta_{ij}^{t}=0\) and \(q_{ij}^{t}=1\), whereas the contact score
decreases exponentially as the separation between the masks increases.

Motion coupling is measured as
\begin{equation}
\kappa_{ij}^{t}
=
r_i^t r_j^t
\exp\left(
-\left\lVert
\mathbf{m}_i^t-\mathbf{m}_j^t
\right\rVert_2
\right),
\label{eq:motion_coupling}
\end{equation}
where \(r_i^t r_j^t\) ensures that motion coupling is evaluated only
when both entities have valid temporal correspondences. A larger value
indicates more similar motion, while unavailable motion contributes
zero coupling.

Based on these predicates, we construct hyperedges using a fixed set
of manipulation templates. Four templates are used:
\begin{equation}
\begin{aligned}
\tau_1 &: \{\mathrm{left\ hand},
             \mathrm{manipulated\ object},
             \mathrm{supporting\ surface}\},\\
\tau_2 &: \{\mathrm{right\ hand},
             \mathrm{manipulated\ object},
             \mathrm{supporting\ surface}\},\\
\tau_3 &: \{\mathrm{left\ hand},
             \mathrm{right\ hand},
             \mathrm{manipulated\ object}\},\\
\tau_4 &: \{\mathrm{active\ hand},
             \mathrm{tool},
             \mathrm{manipulated\ object},
             \mathrm{supporting\ surface}\}.
\end{aligned}
\end{equation}
When at least one hand and the manipulated-object node are available,
the active-hand score is
\begin{equation}
S_{\mathrm{hand}}(h,t)
=
\begin{cases}
\displaystyle
\frac{
I_{h,\mathrm{tool}}^t+
I_{h,\mathrm{obj}}^t
}{2},
&
\text{if a tool is available},\\[3mm]
I_{h,\mathrm{obj}}^t,
&
\text{otherwise}.
\end{cases}
\end{equation}
The active hand is selected as
\begin{equation}
h_t^*
=
\arg\max_{h\in\mathcal{V}_t^{\mathrm{hand}}}
S_{\mathrm{hand}}(h,t).
\end{equation}
When no hand is detected, the corresponding hand-dependent template is
not instantiated. No ground-truth or predicted activity label is used
in this selection.

For candidate hyperedge \(e_j^t\), the predicate vector
\(\mathbf{r}_j^t\) is constructed from the average pairwise contact,
proximity, and motion-coupling values among its member nodes. Let
\begin{equation}
C_j^t
=
\binom{|e_j^t|}{2}
\end{equation}
denote the number of unordered entity pairs within the hyperedge. The
predicate vector is defined as
\begin{equation}
\mathbf{r}_j^t
=
\left[
\frac{1}{C_j^t}
\sum_{\substack{a<b\\v_a^t,v_b^t\in e_j^t}}
q_{ab}^{t};
\frac{1}{C_j^t}
\sum_{\substack{a<b\\v_a^t,v_b^t\in e_j^t}}
\pi_{ab}^{t};
\frac{1}{C_j^t}
\sum_{\substack{a<b\\v_a^t,v_b^t\in e_j^t}}
\kappa_{ab}^{t}
\right].
\end{equation}

The relevance score of the candidate hyperedge is then computed as
\begin{equation}
\alpha_j^t
=
\sigma
\left(
\phi_{\mathrm{score}}
\left(
[\mathbf{r}_j^t;\boldsymbol{\mu}_{\tau(j)}]
\right)
\right),
\end{equation}
where \(\boldsymbol{\mu}_{\tau(j)}\) is a learnable embedding of the
corresponding hyperedge template, \(\phi_{\mathrm{score}}\) is a small
multilayer perceptron, and \(\sigma(\cdot)\) is the sigmoid function.
At each temporal step, up to the top \(K\) valid candidate hyperedges
are retained according to \(\alpha_j^t\). We use \(K=3\), which
limits the number of retained hyperedges. When fewer than \(K\) valid
candidates are available, all valid candidates are retained.

The selected hyperedges define the incidence matrix
\begin{equation}
\mathbf{A}_t
\in
\{0,1\}^{N_t\times M_t},
\end{equation}
where
\begin{equation}
A_{ij}^{t}
=
\begin{cases}
1, & \text{if } v_i^t\in e_j^t,\\
0, & \text{otherwise}.
\end{cases}
\end{equation}
Here, \(N_t\) is the number of entity nodes and \(M_t\) is the number
of selected hyperedges at time \(t\).
We define the set of valid temporal steps as
\begin{equation}
\mathcal{T}_{\mathrm{valid}}
=
\{t\in\{1,\ldots,T\}\mid N_t>0\}.
\end{equation}
If \(N_t=0\), temporal step \(t\) is excluded from the subsequent
temporal-attention normalization. If \(N_t>0\) but no valid candidate
hyperedge can be instantiated, then \(M_t=0\). In this case, no
artificial hyperedge is introduced; hypergraph message passing is
skipped and the temporal representation is constructed using the
available node features together with a zero hyperedge representation.

The initial representation of each hyperedge is obtained by
aggregating the initial node features of its members together with its
predicate vector and template embedding:
\begin{equation}
\mathbf{g}_{j}^{t,0}
=
\phi_{\mathrm{edge}}
\left(
\left[
\mathrm{Pool}
\left(
\{\mathbf{h}_{i}^{t,0}\mid v_i^t\in e_j^t\}
\right);
\mathbf{r}_{j}^{t};
\boldsymbol{\mu}_{\tau(j)}
\right]
\right),
\end{equation}
where \(\mathrm{Pool}(\cdot)\) denotes mean pooling and
\(\phi_{\mathrm{edge}}\) is a projection module.

As an example, in a \emph{cutting} action, the detected entities may
include the right hand, knife, tomato, and cutting board. The
tool-based template forms a hyperedge connecting these four entities.
If the right hand moves together with the knife and the knife remains
close to the tomato over the supporting board, this hyperedge receives
a high relevance score. In contrast, a pairwise graph would represent
the same situation through separate relations, such as hand--knife,
knife--tomato, and tomato--board, without directly encoding the
complete manipulation unit.

\begin{figure}[t]
    \centering
    \includegraphics[width=1\textwidth]{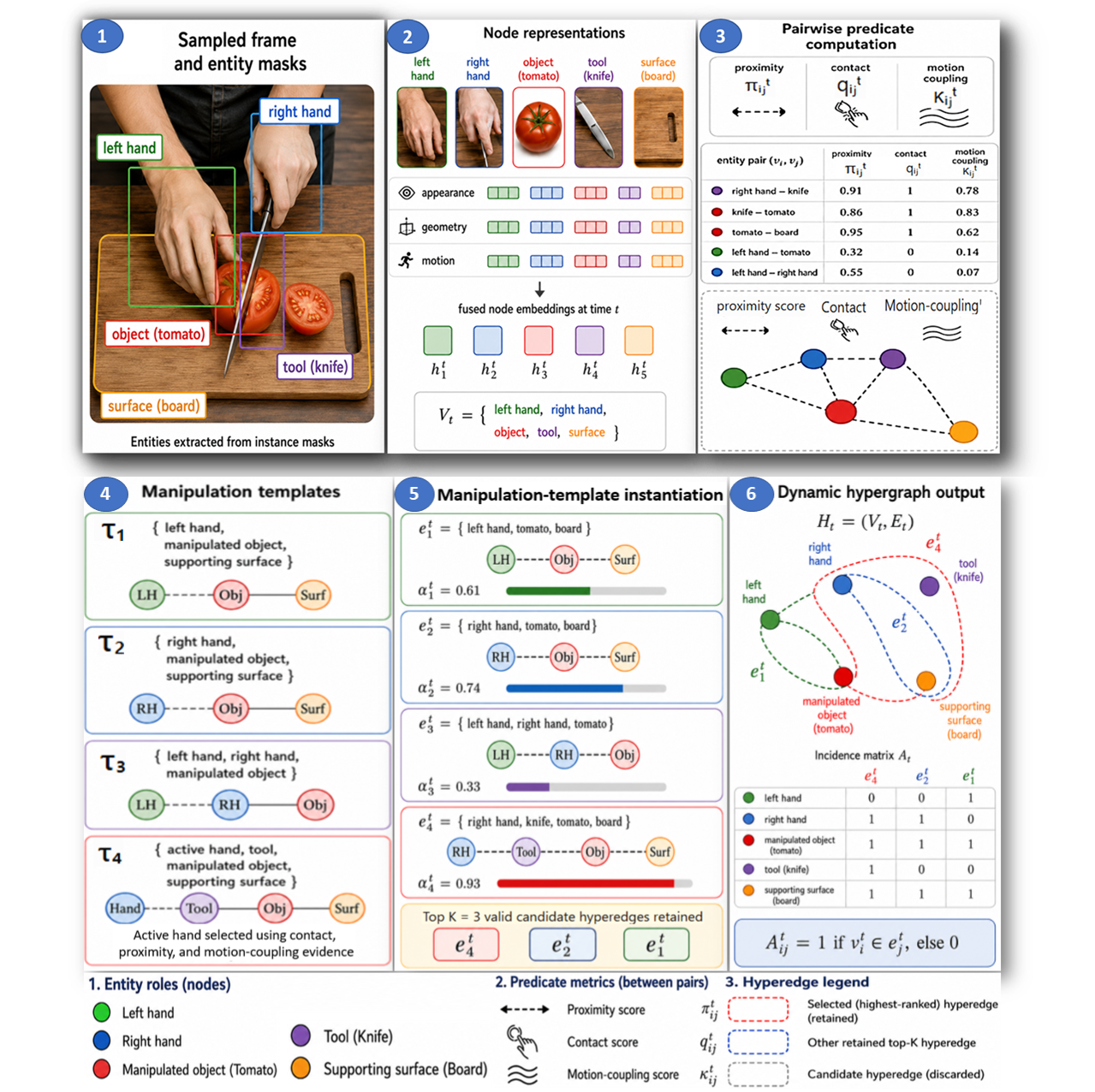}
    \caption{Dynamic manipulation hypergraph construction. Entity masks
are converted into node representations, and proximity
\(\pi_{ij}^t\), contact \(q_{ij}^t\), and motion-coupling
\(\kappa_{ij}^t\) predicates are computed between entity pairs.
Candidate hyperedges are instantiated from the manipulation templates,
and up to the top \(K\) candidates are retained.}
    \label{fig:hypergraph_construction}
\end{figure}

\subsection{Hypergraph Reasoning Network}
\label{subsec:hypergraph_reasoning}

Given the dynamic hypergraph
\(\mathcal{H}_t=(\mathcal{V}_t,\mathcal{E}_t)\), the reasoning network
updates entity and hyperedge representations through higher-order message
passing. Let \(\mathbf{h}_{i}^{t,0}\) denote the initial representation
of node \(v_i^t\), obtained from the entity-representation stage, and let
\(\mathbf{g}_{j}^{t,0}\) denote the initial representation of hyperedge
\(e_j^t\), obtained from the hypergraph-construction stage. The model
applies \(L\) reasoning layers, indexed by
\(\ell=0,\ldots,L-1\), at each temporal step containing at least one
valid hyperedge.

We define the set of valid temporal steps as
\begin{equation}
\mathcal{T}_{\mathrm{valid}}
=
\left\{
t\in\{1,\ldots,T\}\mid N_t>0
\right\}.
\end{equation}
A temporal step with \(N_t=0\) is excluded from temporal-attention
normalization. If \(N_t>0\) but no valid manipulation template can be
instantiated, then \(M_t=0\). In this case, hypergraph message passing
is skipped, and the temporal representation is constructed from the
available node features and a zero hyperedge representation.

For a temporal step with \(M_t>0\), each reasoning layer consists of
node-to-hyperedge aggregation followed by a hyperedge-to-node update.
First, every hyperedge receives information from the nodes participating
in the corresponding manipulation configuration. For hyperedge \(e_j^t\)
at layer \(\ell\), the aggregated node-to-hyperedge message is
\begin{equation}
\tilde{\mathbf{g}}_{j}^{t,\ell}
=
\sum_{i:A_{ij}^{t}=1}
\beta_{ij}^{t,\ell}
\mathbf{W}_{e}^{\ell}
\mathbf{h}_{i}^{t,\ell},
\end{equation}
where \(A_{ij}^{t}\) is the incidence matrix and
\(\beta_{ij}^{t,\ell}\) is an attention weight over the nodes belonging
to hyperedge \(e_j^t\). It is defined as
\begin{equation}
\beta_{ij}^{t,\ell}
=
\frac{
\exp
\left(
\mathbf{u}_{e}^{\top}
\tanh
\left(
\mathbf{W}_{h}^{\ell}\mathbf{h}_{i}^{t,\ell}
+
\mathbf{W}_{g}^{\ell}\mathbf{g}_{j}^{t,\ell}
\right)
\right)
}{
\displaystyle
\sum_{k:A_{kj}^{t}=1}
\exp
\left(
\mathbf{u}_{e}^{\top}
\tanh
\left(
\mathbf{W}_{h}^{\ell}\mathbf{h}_{k}^{t,\ell}
+
\mathbf{W}_{g}^{\ell}\mathbf{g}_{j}^{t,\ell}
\right)
\right)
}.
\end{equation}

The hyperedge representation is then updated using a residual connection:
\begin{equation}
\mathbf{g}_{j}^{t,\ell+1}
=
\mathrm{LN}
\left(
\mathbf{g}_{j}^{t,\ell}
+
\phi_{e}^{\ell}
\left(
[
\mathbf{g}_{j}^{t,\ell};
\tilde{\mathbf{g}}_{j}^{t,\ell}
]
\right)
\right),
\end{equation}
where \(\phi_{e}^{\ell}\) is a multilayer perceptron and
\(\mathrm{LN}(\cdot)\) denotes layer normalization.

After the hyperedges are updated, each node receives information from
the retained hyperedges in which it participates. Let
\begin{equation}
\mathcal{I}_i^t
=
\left\{
j\in\{1,\ldots,M_t\}\mid A_{ij}^t=1
\right\}
\end{equation}
denote the set of retained hyperedges incident on node \(v_i^t\). The
incoming hyperedge-to-node message is defined as
\begin{equation}
\tilde{\mathbf{h}}_{i}^{t,\ell}
=
\begin{cases}
\displaystyle
\sum_{j\in\mathcal{I}_i^t}
\gamma_{ij}^{t,\ell}
\mathbf{W}_{v}^{\ell}
\mathbf{g}_{j}^{t,\ell+1},
&
\mathcal{I}_i^t\neq\varnothing,\\[3mm]
\mathbf{0},
&
\mathcal{I}_i^t=\varnothing.
\end{cases}
\end{equation}

For \(j\in\mathcal{I}_i^t\), the incident-hyperedge attention weight is
\begin{equation}
\gamma_{ij}^{t,\ell}
=
\frac{
\exp
\left(
\mathbf{u}_{v}^{\top}
\tanh
\left(
\mathbf{W}_{v,h}^{\ell}\mathbf{h}_{i}^{t,\ell}
+
\mathbf{W}_{v,g}^{\ell}\mathbf{g}_{j}^{t,\ell+1}
\right)
+
\lambda\alpha_j^t
\right)
}{
\displaystyle
\sum_{r\in\mathcal{I}_i^t}
\exp
\left(
\mathbf{u}_{v}^{\top}
\tanh
\left(
\mathbf{W}_{v,h}^{\ell}\mathbf{h}_{i}^{t,\ell}
+
\mathbf{W}_{v,g}^{\ell}\mathbf{g}_{r}^{t,\ell+1}
\right)
+
\lambda\alpha_r^t
\right)
}.
\end{equation}
Here, \(\alpha_j^t\) is the relevance score of hyperedge \(e_j^t\),
computed during hypergraph construction, and \(\lambda\) controls its
contribution to the attention score. The weight
\(\gamma_{ij}^{t,\ell}\) is evaluated only when
\(\mathcal{I}_i^t\neq\varnothing\); otherwise, the incoming message is
set to the zero vector.

The node representation is updated using
\begin{equation}
\mathbf{h}_{i}^{t,\ell+1}
=
\mathrm{LN}
\left(
\mathbf{h}_{i}^{t,\ell}
+
\phi_{v}^{\ell}
\left(
[
\mathbf{h}_{i}^{t,\ell};
\tilde{\mathbf{h}}_{i}^{t,\ell}
]
\right)
\right),
\end{equation}
where \(\phi_v^\ell\) is a multilayer perceptron. This update allows a
node to incorporate information from the complete manipulation
configurations in which it participates rather than only from independent
pairwise neighbors.

After \(L\) reasoning layers, the node representations are mean-pooled
for each \(t\in\mathcal{T}_{\mathrm{valid}}\):
\begin{equation}
\bar{\mathbf{h}}^{t}
=
\frac{1}{N_t}
\sum_{i=1}^{N_t}
\mathbf{h}_{i}^{t,L}.
\end{equation}
When \(M_t=0\), no reasoning layer is applied and we set
\(\mathbf{h}_{i}^{t,L}=\mathbf{h}_{i}^{t,0}\).

For \(M_t>0\), the normalized relevance weight of hyperedge \(e_j^t\)
is
\begin{equation}
\eta_j^t
=
\frac{\exp(\alpha_j^t)}
{\displaystyle\sum_{r=1}^{M_t}\exp(\alpha_r^t)}.
\end{equation}
The pooled hyperedge representation is defined as
\begin{equation}
\bar{\mathbf{g}}^{t}
=
\begin{cases}
\displaystyle
\sum_{j=1}^{M_t}
\eta_j^t
\mathbf{g}_{j}^{t,L},
&
M_t>0,\\[3mm]
\mathbf{0},
&
M_t=0,
\end{cases}
\end{equation}
where \(\mathbf{0}\) has the same dimensionality as a hyperedge
representation. Thus, temporal steps containing available entities but
no valid hyperedge remain part of the video representation without
requiring an artificial hyperedge.

The representation of valid temporal step \(t\) is
\begin{equation}
\mathbf{z}^{t}
=
\phi_{z}
\left(
[
\bar{\mathbf{h}}^{t};
\bar{\mathbf{g}}^{t}
]
\right),
\qquad
t\in\mathcal{T}_{\mathrm{valid}}.
\end{equation}

Finally, the valid temporal representations are aggregated using temporal
attention. For \(t\in\mathcal{T}_{\mathrm{valid}}\), the temporal
attention weight is
\begin{equation}
\rho_t
=
\frac{
\exp
\left(
\mathbf{u}_{\rho}^{\top}
\tanh
\left(
\mathbf{W}_{\rho}\mathbf{z}^{t}
\right)
\right)
}{
\displaystyle
\sum_{k\in\mathcal{T}_{\mathrm{valid}}}
\exp
\left(
\mathbf{u}_{\rho}^{\top}
\tanh
\left(
\mathbf{W}_{\rho}\mathbf{z}^{k}
\right)
\right)
}.
\end{equation}
For invalid temporal steps, \(t\notin\mathcal{T}_{\mathrm{valid}}\), we
set \(\rho_t=0\). The final video-level representation is
\begin{equation}
\mathbf{z}
=
\sum_{t\in\mathcal{T}_{\mathrm{valid}}}
\rho_t\mathbf{z}^{t}.
\end{equation}

If \(\mathcal{T}_{\mathrm{valid}}=\varnothing\), the final representation
is set to \(\mathbf{z}=\mathbf{0}\). The classification head can then
produce a prediction from its learned bias parameters, while no
ranked hyperedge is returned.

This reasoning process enables information exchange through explicit
multi-entity manipulation units. In a \emph{cutting} action, for example, the representation of the knife is
updated not only from the hand or manipulated object separately, but
from the complete hyperedge connecting the hand, knife, manipulated
object, and cutting board. An ordinary pairwise graph would decompose
the same configuration into several independent relations and would not
directly represent the complete manipulation unit.

\subsection{Activity Prediction and Class-Agnostic Hyperedge Importance}
\label{subsec:prediction_explanation}

The final video-level representation \(\mathbf{z}\), obtained after temporal aggregation, is used for activity classification. A linear prediction head followed by a softmax function maps \(\mathbf{z}\) to a probability distribution over the activity label set \(\mathcal{Y}\):
\begin{equation}
    \mathbf{p}
    =
    \mathrm{softmax}
    \left(
    \mathbf{W}_{c}\mathbf{z}+\mathbf{b}_{c}
    \right),
\end{equation}
where \(\mathbf{W}_{c}\) and \(\mathbf{b}_{c}\) are learnable classification parameters. The predicted activity label is then given by
\begin{equation}
    \hat{y}
    =
    \arg\max_{c \in \mathcal{Y}} p_c .
\end{equation}
During training, the model is optimized using the standard cross-entropy loss,
\begin{equation}
    \mathcal{L}_{\mathrm{cls}}
    =
    -\sum_{c=1}^{|\mathcal{Y}|}
    y_c \log p_c ,
\end{equation}
where \(y_c\) is the one-hot ground-truth label for class \(c\).

In addition to the activity prediction, the model provides a
class-agnostic hyperedge-importance score. For retained hyperedge
\(e_j^t\), the score is defined as
\begin{equation}
\omega_j^t
=
\rho_t\eta_j^t,
\label{eq:hyperedge_importance}
\end{equation}
where \(\rho_t\) is the temporal-attention weight and \(\eta_j^t\) is
the normalized within-step relevance weight of hyperedge \(e_j^t\).
The score ranks the relational patterns emphasized by the model but is
not a causal attribution to the predicted class.

Let the total number of retained hyperedges in the clip be
\begin{equation}
M_{\mathrm{tot}}
=
\sum_{t\in\mathcal{T}_{\mathrm{valid}}}M_t.
\end{equation}
The ranked hyperedge indices are defined as
\begin{equation}
\mathcal{I}_{\mathrm{exp}}
=
\begin{cases}
\displaystyle
\operatorname{TopIndex}_{\min(R,M_{\mathrm{tot}})}
\left(
\left\{
(\omega_j^t,t,j)
\mid
t\in\mathcal{T}_{\mathrm{valid}},
\;j=1,\ldots,M_t
\right\}
\right),
&
M_{\mathrm{tot}}>0,\\[4mm]
\varnothing,
&
M_{\mathrm{tot}}=0.
\end{cases}
\end{equation}
The reported hyperedge set is
\begin{equation}
\mathcal{E}_{\mathrm{exp}}
=
\left\{
e_j^t
\mid
(t,j)\in\mathcal{I}_{\mathrm{exp}}
\right\}.
\end{equation}
Here, \(\operatorname{TopIndex}\) returns the temporal-step and
hyperedge indices associated with the largest importance scores.

For visualization only, the displayed scores are divided by the
maximum hyperedge-importance score within each clip. This normalization
does not affect hyperedge ranking. In failure cases, the ranked
hyperedges may suggest missing entities, weak relational evidence, or
incorrect temporal focus, but they do not establish the cause of the
error.

\subsection*{Algorithmic Summary}

The complete flow of the proposed method is summarized in
Algorithm~\ref{alg:dmh}. The algorithm shows how candidate entities are assigned to functional
roles, represented as nodes, organized into dynamic manipulation
hyperedges, and processed to obtain the predicted activity and
top-ranked hyperedges.

\begin{algorithm}[H]
\caption{Dynamic Manipulation Hypergraph Inference}
\label{alg:dmh}
\scriptsize
\algrenewcommand\algorithmicindent{0.9em}
\begin{algorithmic}[1]

\Require Video clip \(X=\{x_t\}_{t=1}^{T}\), limits \(K,L,R\)
\Ensure Predicted activity \(\hat{y}\) and ranked hyperedges
\(\mathcal{E}_{\mathrm{exp}}\)

\Statex \textit{Track-level entity processing}

\For{\(t=1,\ldots,T\)}
    \State Extract candidate masks \(\mathcal{C}_t\) and compute their
    appearance and spatial descriptors.
\EndFor

\State Associate candidates across sampled frames to form entity tracks.

\For{\(t=1,\ldots,T\)}
    \State Compute motion descriptors and validity indicators \(r_i^t\).
    \State Compute \(d_{ij}^t,\delta_{ij}^t,\pi_{ij}^t,q_{ij}^t,
    \kappa_{ij}^t\), and \(I_{ij}^t\) for observed candidate pairs.
\EndFor

\State Compute track persistence and temporally averaged role scores.
\State Assign selected tracks to the left-hand, right-hand,
manipulated-object, tool, and supporting-surface roles.
\State \(\mathcal{T}_{\mathrm{valid}}\leftarrow\varnothing\)

\Statex \textit{Dynamic hypergraph construction and reasoning}

\For{\(t=1,\ldots,T\)}

    \State Form \(\mathcal{V}_t=\{v_i^t\}_{i=1}^{N_t}\) from the
    assigned tracks observed at \(t\).

    \If{\(N_t=0\)}
        \State \textbf{continue}
    \EndIf

    \State
    \(\mathcal{T}_{\mathrm{valid}}
    \leftarrow\mathcal{T}_{\mathrm{valid}}\cup\{t\}\)

    \State Initialize each node as
    \(\mathbf{h}_i^{t,0}
    \leftarrow
    \phi_{\mathrm{ent}}
    ([\mathbf{a}_i^t;\mathbf{b}_i^t;
    \mathbf{m}_i^t;\mathbf{s}_i^t])\).

    \If{a hand and the manipulated object are available}
        \State Select the active hand
        \(h_t^*\leftarrow
        \arg\max_{h\in\mathcal{V}_t^{\mathrm{hand}}}
        S_{\mathrm{hand}}(h,t)\).
    \EndIf

    \State Instantiate
    \(\widetilde{\mathcal{E}}_t
    =\{\widetilde e_j^t\}_{j=1}^{\widetilde M_t}\)
    from the available manipulation templates.

    \For{\(j=1,\ldots,\widetilde M_t\)}
        \State Construct \(\widetilde{\mathbf r}_j^t\) and compute
        \(\widetilde\alpha_j^t
        \leftarrow
        \sigma(\phi_{\mathrm{score}}
        ([\widetilde{\mathbf r}_j^t;
        \boldsymbol{\mu}_{\tau(j)}]))\).
    \EndFor

    \State Set \(M_t\leftarrow\min(K,\widetilde M_t)\).

    \If{\(M_t>0\)}

        \State Select
        \[
        (j_1,\ldots,j_{M_t})
        \leftarrow
        \operatorname{TopIndex}_{M_t}
        (\{\widetilde\alpha_j^t\}_{j=1}^{\widetilde M_t}).
        \]

        \State Retain and consecutively re-index
        \(\widetilde e_{j_r}^t\), \(\widetilde{\mathbf r}_{j_r}^t\),
        and \(\widetilde\alpha_{j_r}^t\) as
        \(e_r^t\), \(\mathbf r_r^t\), and \(\alpha_r^t\).

        \State Form
        \(\mathcal{E}_t=\{e_r^t\}_{r=1}^{M_t}\),
        construct \(\mathbf A_t\), and initialize
        \(\mathbf g_r^{t,0}\).

        \For{\(\ell=0,\ldots,L-1\)}
            \State Compute \(\beta_{ir}^{t,\ell}\) and apply the
            node-to-hyperedge update.
            \State Compute relevance-biased
            \(\gamma_{ir}^{t,\ell}\) and apply the hyperedge-to-node
            update.
            \State Use a zero incoming message for each node with no
            incident retained hyperedge.
        \EndFor

        \State Compute
        \(\eta_r^t
        \leftarrow
        \exp(\alpha_r^t)/
        \sum_{s=1}^{M_t}\exp(\alpha_s^t)\).

        \State Compute
        \(\bar{\mathbf g}^t
        \leftarrow
        \sum_{r=1}^{M_t}\eta_r^t\mathbf g_r^{t,L}\).

    \Else

        \State Set
        \(\mathcal{E}_t\leftarrow\varnothing\),
        \(\mathbf h_i^{t,L}\leftarrow\mathbf h_i^{t,0}\), and
        \(\bar{\mathbf g}^t\leftarrow\mathbf0\).

    \EndIf

    \State Compute
    \(\bar{\mathbf h}^t
    \leftarrow
    N_t^{-1}\sum_{i=1}^{N_t}\mathbf h_i^{t,L}\).

    \State Compute
    \(\mathbf z^t
    \leftarrow
    \phi_z([\bar{\mathbf h}^t;\bar{\mathbf g}^t])\).

\EndFor

\Statex \textit{Prediction and hyperedge ranking}

\If{\(\mathcal{T}_{\mathrm{valid}}=\varnothing\)}
    \State Set \(\mathbf z\leftarrow\mathbf0\).
\Else
    \State Compute temporal-attention weights
    \(\{\rho_t:t\in\mathcal{T}_{\mathrm{valid}}\}\).
    \State Set
    \(\mathbf z
    \leftarrow
    \sum_{t\in\mathcal{T}_{\mathrm{valid}}}
    \rho_t\mathbf z^t\).
\EndIf

\State Compute
\(\mathbf p\leftarrow
\operatorname{softmax}(\mathbf W_c\mathbf z+\mathbf b_c)\)
and
\(\hat y\leftarrow\arg\max_{c\in\mathcal Y}p_c\).

\If{\(\sum_{t\in\mathcal{T}_{\mathrm{valid}}}M_t=0\)}
    \State Set \(\mathcal{E}_{\mathrm{exp}}\leftarrow\varnothing\).
\Else
    \State Compute
    \(\omega_j^t\leftarrow\rho_t\eta_j^t\)
    for every retained hyperedge.

    \State Select
    \[
    \mathcal I_{\mathrm{exp}}
    \leftarrow
    \operatorname{TopIndex}_{
\min\left(
R,\sum_{t\in\mathcal{T}_{\mathrm{valid}}}M_t
\right)}
\left(
\left\{
(\omega_j^t,t,j)
\mid
t\in\mathcal{T}_{\mathrm{valid}},
\;j=1,\ldots,M_t
\right\}
\right).
    \]

    \State Set
    \(\mathcal E_{\mathrm{exp}}
    \leftarrow
    \{e_j^t:(t,j)\in\mathcal I_{\mathrm{exp}}\}\).
\EndIf

\State \Return \(\hat y\) and \(\mathcal E_{\mathrm{exp}}\)

\end{algorithmic}
\end{algorithm}
\FloatBarrier
\section{Experiments and Results}
\label{sec:experiments_results}

This section presents the experimental evaluation of the proposed dynamic
manipulation hypergraph model. We first describe the datasets, selected action categories, and evaluation protocol. We then provide the implementation details, including entity extraction, feature construction, model configuration, and training settings. Next, we introduce the comparison baselines and evaluation metrics. The main quantitative results are followed by ablation studies and a family-wise analysis of higher-order manipulation actions. Finally, qualitative examples illustrate the relational evidence captured by the learned hyperedges and highlight representative failure cases.

\subsection{Datasets and Experimental Protocol}
\label{subsec:dataset_protocol}

Quantitative experiments are conducted on
EPIC-KITCHENS-100/VISOR
\cite{Damen_2022_IJCV_EPIC100,Darkhalil_2022_NeurIPS_VISOR}
and Assembly101~\cite{Sener_2022_CVPR_Assembly101}.
ARCTIC~\cite{Fan_2023_CVPR_ARCTIC} supports the qualitative
contact-oriented analysis.

\paragraph{Class selection.}
Activity classes were selected before training according to whether
their definitions involved hand--object--surface, bimanual
hand--object, or hand--tool--object--surface interactions. Selection
was based on the activity definitions rather than on successful entity
detection; therefore, clips with missing entities remained in the
evaluation. A predefined subset was organized into surface-dependent,
tool-mediated, bimanual, and contact- or state-change families. All
classes contribute to accuracy and macro-\(F_1\), whereas the
predefined subset contributes to HO-\(F_1\).

\paragraph{EPIC-KITCHENS-100/VISOR.}
EPIC-KITCHENS-100 provides the temporal action segments and activity
labels, while VISOR provides the hand and active-object masks used as
annotation-assisted entity inputs. Supporting surfaces are obtained using the
procedure described in
Section~\ref{subsec:implementation_details}. A fixed internal split was
constructed from the publicly labelled clips with the required VISOR
annotations. Partitioning was performed by source video before class
selection. The resulting subset contains 8,920 clips from 14 classes:
6,240 for training, 1,320 for validation, and 1,360 for testing.

\paragraph{Assembly101.}
The egocentric streams and annotated fine-grained action segments of
Assembly101 are used. The original data partitions were retained, and
class filtering was applied independently within each partition. The
resulting subset contains 26,600 clips from 20 classes: 18,600 for
training, 3,900 for validation, and 4,100 for testing. Hands,
manipulated components, tools, and supporting surfaces are obtained
using the procedure described in
Section~\ref{subsec:implementation_details}.

\paragraph{ARCTIC.}
ARCTIC is used to compare normalized hyperedge-importance scores with
annotated bimanual-contact intervals. The frozen
EPIC-KITCHENS-100/VISOR checkpoint selected according to validation
macro-\(F_1\) is applied without adaptation. Contact annotations are
accessed only after inference and are not used for training or model
selection. No aggregate metric is calculated on ARCTIC.

\paragraph{Evaluation protocol.}
The validation sets are used for hyperparameter selection, threshold
calibration, and checkpoint selection, while the test sets are reserved
for final evaluation. Each action segment is uniformly sampled into
\(T=16\) temporal steps. All methods use the same clips, temporal
samples, and partitions. The controlled entity-based, graph-based, and
hypergraph-based models additionally use identical entity regions,
appearance features, and node descriptors. Dataset composition is
summarized in Table~\ref{tab:datasets_protocol}.

\begin{table}[t]
\centering
\caption{Composition and evaluation role of the datasets. Clip counts refer
to the final manipulation subsets obtained using the predefined
class-selection protocol.}
\label{tab:datasets_protocol}
\renewcommand{\arraystretch}{1.16}
\small
\begin{tabularx}{\textwidth}{p{2.5cm} C C C C X}
\hline
\textbf{Dataset} &
\textbf{Train} &
\textbf{Val.} &
\textbf{Test} &
\textbf{Classes} &
\textbf{Evaluation role} \\
\hline

EPIC-KITCHENS-100/VISOR &
6,240 &
1,320 &
1,360 &
14 &
Daily egocentric manipulation recognition \\
\hline

Assembly101 &
18,600 &
3,900 &
4,100 &
20 &
Procedural bimanual and tool-mediated recognition \\
\hline

ARCTIC &
-- &
-- &
-- &
-- &
Contact-oriented temporal analysis only \\
\hline
\end{tabularx}
\end{table}

\subsection{Implementation Details}
\label{subsec:implementation_details}

\paragraph{Temporal sampling and entity extraction.}
Each action clip is uniformly sampled into \(T=16\) temporal steps.
Entity detection and segmentation are performed using frozen pretrained
models. Grounding DINO with the Swin-T checkpoint is used for
text-conditioned candidate detection~\cite{Liu_2024_ECCV_GroundingDINO},
and SAM~2 with the Hiera-Large checkpoint is used to obtain and
temporally propagate instance masks~\cite{Ravi_2025_ICLR_SAM2}. The
models are not fine-tuned on the activity-recognition test sets.

The candidate vocabulary contains hand, object, tool, component, table,
workbench, board, and supporting-surface categories, together with
dataset-specific object categories observed in the training partition.
The vocabulary is fixed before validation and test evaluation. In
particular, the ground-truth activity label of an evaluated clip is
never used as a detection prompt. Grounding DINO detections with a
confidence below \(0.35\) are removed, and non-maximum suppression is
applied with an intersection-over-union threshold of \(0.50\).

\paragraph{Dataset-specific entity inputs.}
For EPIC-KITCHENS-100/VISOR, the available VISOR hand and active-object
masks are used as annotation-assisted entity inputs. Tools and
supporting surfaces not explicitly included in the VISOR annotations
are obtained using the fixed Grounding DINO--SAM~2 pipeline. Therefore,
the EPIC-KITCHENS-100/VISOR experiment should be interpreted as an
annotation-assisted evaluation of relational reasoning rather than as
a completely automatic perception setting.

For Assembly101, the official three-dimensional hand poses are
projected into the selected egocentric view using the provided camera
calibration parameters. A tight bounding box is constructed around the
visible joints and enlarged by \(15\%\) in each spatial direction.
These boxes are used as prompts for SAM~2 to obtain left- and right-hand
masks. Manipulated components, tools, and supporting work surfaces are
generated using the same Grounding DINO--SAM~2 pipeline employed for
non-annotated EPIC-KITCHENS entities.

The Assembly101 fine-grained action annotations are used only to define
the temporal clip boundaries and classification targets. They are not
used to determine the principal component, tool, surface, or active
hand. The principal component is selected using the label-independent
contact, proximity, motion-coupling, confidence, and temporal-
persistence criteria defined in
Section~\ref{subsec:entity_representation}. The tool and supporting
surface roles are assigned using the corresponding label-independent
role-selection rules. Clips with missing entities remain in the
evaluation set.

For ARCTIC, hand and object masks are generated using the same frozen
extraction pipeline. Ground-truth contact annotations are accessed only after inference to qualitatively illustrate the temporal correspondence between annotated
contact intervals and the predicted hyperedge-importance scores. They are not used during entity selection, hypergraph construction, training, checkpoint selection, or activity prediction.

\paragraph{Temporal association and feature extraction.}
SAM~2 propagates each prompted entity mask across the sampled frames.
A propagated track is retained as a principal entity candidate when it
is observed in at least three of the 16 temporal steps. Shorter tracks
are treated as transient detections. When several candidates satisfy
the same functional role, the candidate with the highest temporally
averaged role-assignment score is selected. Missing roles are omitted
at the corresponding temporal steps, but the video clip is not
discarded.

Each selected mask is converted into a tight bounding box, and its
image crop is resized to \(224\times224\) pixels. Appearance features
are extracted using an ImageNet-pretrained ResNet-50 backbone. The
backbone remains frozen during training. The same masks, tracks, entity
crops, and appearance features are supplied to all entity-based,
pairwise-graph, static-hypergraph, and dynamic-hypergraph models.

\paragraph{Hypergraph configuration.}
Node and hyperedge representations have a dimensionality of \(256\).
The entity-projection modules contain two fully connected layers with
ReLU activation and dropout of \(0.20\). The hypergraph reasoning
network contains \(L=3\) layers. At each temporal step, at most \(K=3\)
valid hyperedges are retained. If fewer than three valid candidates
are available, all valid candidates are retained; no artificial
hyperedge is introduced.

Spatial distances are normalized by the frame diagonal. The proximity
and contact-distance scales are set to \(\tau_d=0.20\) and
\(\tau_c=0.01\), respectively. The relevance coefficient in the
hyperedge-to-node attention was fixed to \(\lambda=1\).
For qualitative reporting, the highest-ranked hyperedge was retained,
corresponding to \(R=1\).

\paragraph{Optimization and model selection.}
The model is trained using AdamW with an initial learning rate of
\(3\times10^{-4}\), weight decay of \(1\times10^{-4}\), and batch size
of \(16\) for 60 epochs. A cosine-annealing learning-rate schedule is
used after a linear warm-up over the first five epochs. The checkpoint
with the highest validation macro-\(F_1\) is selected independently for
each run.

All experiments are conducted on one NVIDIA RTX 4090 GPU using random
seeds 13, 37, and 73. Test results are reported as the mean $\pm$ standard deviation across the three runs. The validation partition is used for
checkpoint selection, hyperparameter tuning, and threshold calibration;
the held-out test partition is used only for final evaluation.

\subsection{Baselines and Evaluation Metrics}
\label{subsec:baselines_metrics}

We compare the proposed method with video-level, entity-based, graph-based, and hypergraph-based baselines. All relational methods use the same action segments, dataset splits, temporal sampling, visual backbone, and entity features whenever applicable.

\paragraph{Video-level baselines.}
VideoMAE~V2~\cite{Wang_2023_CVPR_VideoMAEv2} and InternVideo2~\cite{Wang_2024_ECCV_InternVideo2} classify each clip from global spatio-temporal features without explicitly representing manipulation entities or their relations.

\paragraph{Entity and graph baselines.}
Our contact-aware entity baseline, adapted from Shiota et al.~\cite{Shiota_2024_WACV_ContactState}, combines pooled entity features with the
same proximity, contact, and motion-coupling predicates used by the proposed
method, but does not construct a graph or hypergraph. We additionally construct a matched pairwise graph baseline using the same nodes, predicates, temporal sampling, and reasoning depth as the proposed method. Each higher-order candidate configuration is replaced by ordinary pairwise connections among its participating entities. This baseline directly evaluates whether higher-order hyperedges provide an advantage over pairwise relations.

We also compare with our hierarchical graph model for bimanual action
recognition~\cite{Ziaeetabar_2024_IEEEAccess_HierarchicalBimanual} and
a visual-only adaptation of the graph-reasoning model in
\cite{Ziaeetabar_2025_IEEEAccess_AdaptiveMultimodal}. Both are adapted
to the same evaluated clips and visual backbone.

\paragraph{Static hypergraph baseline.}
The static hypergraph uses the same entity roles, manipulation
templates, node features, and reasoning layers as the proposed model.
Template identities are selected once using clip-level averaged
predicates and are not re-ranked over time. At temporal steps where a
required role is absent, the corresponding template is masked.
Comparison with this baseline evaluates the contribution of per-step
dynamic template selection.

Table~\ref{tab:baselines} summarizes the input representation,
initialization, training protocol, and random seeds used for each
evaluated method.

\begin{table}[!t]
\centering
\caption{Implementation configuration of the evaluated baselines.
All methods use the same dataset partitions and temporal sampling.}
\label{tab:baselines}
\renewcommand{\arraystretch}{1.15}
\scriptsize
\begin{tabularx}{\textwidth}{
p{2.6cm}
p{2.4cm}
p{2.7cm}
X
p{1.1cm}}
\hline
\textbf{Method} &
\textbf{Input} &
\textbf{Backbone} &
\textbf{Training and adaptation} &
\textbf{Seeds} \\
\hline

VideoMAE~V2~\cite{Wang_2023_CVPR_VideoMAEv2} &
16 global RGB frames &
Official pretrained VideoMAE~V2 encoder &
Frozen video encoder; classification head trained on each selected dataset subset &
13, 37, 73 \\
\hline

InternVideo2~\cite{Wang_2024_ECCV_InternVideo2} &
16 global RGB frames &
Official pretrained InternVideo2 encoder &
Frozen video encoder; classification head trained on each selected dataset subset &
13, 37, 73 \\
\hline

Contact-aware entity model~\cite{Shiota_2024_WACV_ContactState} &
16 sets of entity crops and predicates &
Frozen ImageNet-pretrained ResNet-50 &
Entity-fusion and classification modules retrained using the common protocol &
13, 37, 73 \\
\hline

Matched pairwise graph &
16 pairwise entity graphs &
Frozen ImageNet-pretrained ResNet-50 &
Graph layers and classifier trained with the same node dimensions and reasoning depth as the proposed model &
13, 37, 73 \\
\hline

Hierarchical graph model~\cite{Ziaeetabar_2024_IEEEAccess_HierarchicalBimanual} &
16 hierarchical entity graphs &
Frozen ImageNet-pretrained ResNet-50 &
Adapted and retrained on the same clips using the common entity features &
13, 37, 73 \\
\hline

Adaptive graph reasoning (visual-only adaptation)~\cite{Ziaeetabar_2025_IEEEAccess_AdaptiveMultimodal} &
16 adaptive entity graphs &
Frozen ImageNet-pretrained ResNet-50 &
Adapted and retrained on the same clips using the common entity features &
13, 37, 73 \\
\hline

Static hypergraph &
16 fixed-template hypergraphs &
Frozen ImageNet-pretrained ResNet-50 &
Templates selected once using clip-averaged predicates &
13, 37, 73 \\
\hline

Proposed method &
16 dynamic hypergraphs &
Frozen ImageNet-pretrained ResNet-50 &
Dynamic hypergraph layers and classifier trained using the common optimization protocol &
13, 37, 73 \\
\hline
\end{tabularx}
\end{table}

For the entity-based and relational models, the common optimization
protocol uses AdamW with an initial learning rate of
\(3\times10^{-4}\), weight decay of \(1\times10^{-4}\), batch size
16, and 60 training epochs. Model selection is performed independently
for each method and random seed according to validation macro-\(F_1\).
All entity-based, graph-based, and hypergraph-based methods use the
same extracted entity regions and frozen ResNet-50 appearance features.

Performance is evaluated using top-1 accuracy, macro-\(F_1\), and
higher-order action \(F_1\) (HO-\(F_1\)). Macro-\(F_1\) assigns equal
importance to every activity class and therefore reduces the influence
of class imbalance. Let \(\mathcal{Y}_{\mathrm{HO}}\) denote the
predefined union of surface-dependent, tool-mediated, bimanual, and
contact- or state-change activity classes. HO-\(F_1\) is defined as
\begin{equation}
\mathrm{HO}\text{-}F_1
=
\frac{1}{|\mathcal{Y}_{\mathrm{HO}}|}
\sum_{c\in\mathcal{Y}_{\mathrm{HO}}}F_{1,c},
\end{equation}
where \(F_{1,c}\) is the class-wise \(F_1\) score for class \(c\).
A class belonging to multiple manipulation families is counted only
once in \(\mathcal{Y}_{\mathrm{HO}}\). The higher-order class set is
fixed before test evaluation and is used only for analysis; training
uses the original dataset activity labels. Family-wise \(F_1\) scores are
additionally reported to identify the manipulation categories that
benefit most from higher-order reasoning. 

\FloatBarrier

\subsection{Main Quantitative Results}
\label{subsec:main_results}

Table~\ref{tab:main_results} compares the proposed method with
video-level, entity-based, graph-based, and static-hypergraph baselines
under the common evaluation protocol described in
Section~\ref{subsec:dataset_protocol}. Quantitative results are reported on EPIC-KITCHENS-100/VISOR and Assembly101. ARCTIC is excluded from this comparison because it is used only for supporting contact-oriented and qualitative analysis.

\begin{table}[!htbp]
\centering
\caption{Main quantitative comparison under the common evaluation
protocol. Values are reported as mean $\pm$ standard deviation across
three random seeds.}
\label{tab:main_results}
\renewcommand{\arraystretch}{1.14}
\small
\begin{tabularx}{\textwidth}{p{2.6cm} Y C C C}
\hline
\textbf{Dataset} &
\textbf{Method} &
\textbf{Acc.} &
\textbf{Macro-F1} &
\textbf{HO-F1} \\
\hline

\multirow{8}{2.6cm}{EPIC-KITCHENS-100/VISOR}
& VideoMAE~V2
& $57.9 \pm 0.8$
& $53.8 \pm 0.9$
& $49.6 \pm 1.0$ \\

& InternVideo2
& $60.7 \pm 0.7$
& $56.2 \pm 0.8$
& $52.1 \pm 0.9$ \\

& Contact-aware entity model
& $61.8 \pm 0.6$
& $57.7 \pm 0.7$
& $55.3 \pm 0.8$ \\

& Matched pairwise graph
& $63.9 \pm 0.5$
& $60.0 \pm 0.6$
& $58.2 \pm 0.8$ \\

& Hierarchical graph model
& $62.6 \pm 0.7$
& $58.8 \pm 0.7$
& $56.4 \pm 0.9$ \\

& Adaptive graph reasoning (visual-only adaptation)
& $65.1 \pm 0.6$
& $61.4 \pm 0.6$
& $59.0 \pm 0.7$ \\

& Static hypergraph
& $65.7 \pm 0.5$
& $62.0 \pm 0.6$
& $60.7 \pm 0.7$ \\

& Proposed method
& $\mathbf{68.2 \pm 0.4}$
& $\mathbf{64.6 \pm 0.5}$
& $\mathbf{65.1 \pm 0.6}$ \\
\hline

\multirow{8}{2.6cm}{Assembly101}
& VideoMAE~V2
& $43.8 \pm 0.9$
& $39.5 \pm 1.0$
& $37.2 \pm 1.1$ \\

& InternVideo2
& $47.1 \pm 0.8$
& $42.9 \pm 0.9$
& $40.8 \pm 1.0$ \\

& Contact-aware entity model
& $46.0 \pm 0.8$
& $42.3 \pm 0.8$
& $41.7 \pm 0.9$ \\

& Matched pairwise graph
& $50.2 \pm 0.6$
& $46.8 \pm 0.7$
& $45.5 \pm 0.8$ \\

& Hierarchical graph model
& $49.0 \pm 0.7$
& $45.6 \pm 0.8$
& $44.8 \pm 0.9$ \\

& Adaptive graph reasoning (visual-only adaptation)
& $51.4 \pm 0.7$
& $48.2 \pm 0.7$
& $46.7 \pm 0.8$ \\

& Static hypergraph
& $52.0 \pm 0.6$
& $48.8 \pm 0.7$
& $49.2 \pm 0.7$ \\

& Proposed method
& $\mathbf{55.3 \pm 0.5}$
& $\mathbf{52.1 \pm 0.6}$
& $\mathbf{55.0 \pm 0.7}$ \\
\hline
\end{tabularx}
\end{table}

The proposed method achieves the highest mean performance under the
evaluated protocol, with the matched pairwise graph and static
hypergraph providing the principal controlled comparisons. Compared with InternVideo2, the strongest video-only baseline, it improves HO-F1 by 13.0 percentage points on EPIC-KITCHENS-100/VISOR and 14.2 points on Assembly101. The larger mean difference in HO-F1 indicates a greater performance advantage on the predefined higher-order class subset. 
Comparisons with VideoMAE~V2 and InternVideo2 should be interpreted as
contextual rather than strictly controlled, because their pretrained video encoders remain frozen and only their classification heads are trained.
The matched pairwise graph and static hypergraph therefore constitute the primary controlled baselines, as they use identical entity features, temporal sampling, reasoning depth, and optimization settings. 
Replacing pairwise edges with higher-order hyperedges improves accuracy by 4.3 points on EPIC-KITCHENS-100/VISOR and 5.1 points on Assembly101. The corresponding mean HO-F1 differences are 6.9 and 9.5 points,
respectively, showing that the proposed higher-order formulation
achieves higher mean scores than the matched pairwise formulation under
the evaluated protocol.

The proposed method also outperforms the static hypergraph by 2.5 accuracy points and 4.4 HO-F1 points on EPIC-KITCHENS-100/VISOR, and by 3.3 accuracy points and 5.8 HO-F1 points on Assembly101. The higher mean scores relative to the static baseline suggest an additional benefit from per-step dynamic template selection under the
evaluated protocol.

For completeness, Table~\ref{tab:relative_gains} summarizes the gains over the three most informative baselines.

\begin{table}[!htbp]
\centering
\caption{Improvement of the proposed method over selected baselines.}
\label{tab:relative_gains}
\renewcommand{\arraystretch}{1.14}
\small
\begin{tabularx}{\textwidth}{p{2.7cm} Y C C C}
\hline
\textbf{Dataset} &
\textbf{Comparison} &
\textbf{$\Delta$ Acc.} &
\textbf{$\Delta$ Macro-F1} &
\textbf{$\Delta$ HO-F1} \\
\hline

EPIC-KITCHENS-100/VISOR &
Proposed $-$ strongest video-only baseline &
+7.5 & +8.4 & +13.0 \\

EPIC-KITCHENS-100/VISOR &
Proposed $-$ matched pairwise graph &
+4.3 & +4.6 & +6.9 \\

EPIC-KITCHENS-100/VISOR &
Proposed $-$ static hypergraph &
+2.5 & +2.6 & +4.4 \\
\hline

Assembly101 &
Proposed $-$ strongest video-only baseline &
+8.2 & +9.2 & +14.2 \\

Assembly101 &
Proposed $-$ matched pairwise graph &
+5.1 & +5.3 & +9.5 \\

Assembly101 &
Proposed $-$ static hypergraph &
+3.3 & +3.3 & +5.8 \\
\hline
\end{tabularx}
\end{table}

\FloatBarrier

\subsection{Ablation Study}
\label{subsec:ablation}

We evaluate the contribution of the main model components on EPIC-KITCHENS-100/VISOR. Each ablation modifies one component while retaining the same entity inputs, visual features, training settings, and evaluation protocol. We examine the contact and motion-coupling predicates, semantic-role embeddings, learned hyperedge weighting, temporal attention, time-varying hypergraph construction, and higher-order relational representation.

In the uniform hyperedge-pooling variant, the retained hyperedges are
pooled using
\[
\eta_j^t
=
\frac{1}{M_t},
\qquad
j=1,\ldots,M_t,
\]
while all other uses of \(\alpha_j^t\) remain unchanged.

\begin{table}[!htbp]
\centering
\caption{Ablation study on EPIC-KITCHENS-100/VISOR. Values are reported
as mean $\pm$ standard deviation across three random seeds.}
\label{tab:ablation}
\renewcommand{\arraystretch}{1.16}
\small
\begin{tabularx}{\textwidth}{Y C C C}
\hline
\textbf{Model variant} &
\textbf{Acc.} &
\textbf{Macro-F1} &
\textbf{HO-F1} \\
\hline

Without contact predicate
& $66.4 \pm 0.6$
& $62.7 \pm 0.7$
& $61.6 \pm 0.8$ \\

Without motion coupling
& $66.9 \pm 0.5$
& $63.2 \pm 0.6$
& $62.4 \pm 0.7$ \\

Without semantic-role embedding
& $67.3 \pm 0.5$
& $63.8 \pm 0.6$
& $63.2 \pm 0.7$ \\

Uniform hyperedge pooling
& $66.6 \pm 0.6$
& $62.9 \pm 0.7$
& $61.9 \pm 0.8$ \\

Mean temporal pooling
& $66.8 \pm 0.5$
& $63.1 \pm 0.6$
& $62.3 \pm 0.7$ \\

Static hypergraph
& $65.7 \pm 0.5$
& $62.0 \pm 0.6$
& $60.7 \pm 0.7$ \\

Matched pairwise graph
& $63.9 \pm 0.5$
& $60.0 \pm 0.6$
& $58.2 \pm 0.8$ \\

\textbf{Full proposed model}
& $\mathbf{68.2 \pm 0.4}$
& $\mathbf{64.6 \pm 0.5}$
& $\mathbf{65.1 \pm 0.6}$ \\
\hline
\end{tabularx}
\end{table}

The matched pairwise graph produces the largest performance reduction, decreasing accuracy by 4.3 points and HO-F1 by 6.9 points relative to the full model. This result supports the use of higher-order hyperedges for representing complete manipulation configurations rather than decomposing them into independent pairwise relations.

The static hypergraph also performs below the full model, with reductions of 2.5 points in accuracy and 4.4 points in HO-F1. This indicates that constructing the hypergraph from time-varying entity relations contributes beyond the use of higher-order structure alone.

Removing contact information or motion coupling, or replacing
relevance-weighted hyperedge pooling with uniform pooling, causes a
larger reduction in HO-\(F_1\) than in overall accuracy. Contact
information and relevance-weighted pooling are particularly important
for identifying the active multi-entity configuration.

\subsection{Analysis of Higher-Order Manipulation Classes}
\label{subsec:higher_order_analysis}

To identify the action types that benefit most from higher-order reasoning, we report family-wise F1 scores for the manipulation categories defined in Section~\ref{subsec:dataset_protocol}. The analysis is restricted to EPIC-KITCHENS-100/VISOR and Assembly101, which are used for quantitative recognition experiments. ARCTIC is reserved for qualitative contact-oriented analysis.

Table~\ref{tab:higher_order_analysis} compares the matched pairwise graph, static hypergraph, and proposed method. The matched pairwise graph uses the same entities, predicates, temporal sampling, and reasoning depth as the proposed model, but represents each manipulation configuration through ordinary pairwise edges. The static hypergraph retains higher-order relations but uses fixed clip-level template selection.

\begin{table}[!htbp]
\centering
\caption{Family-wise \(F_1\) scores for higher-order manipulation
classes. Gain denotes the improvement of the proposed method over the
matched pairwise graph. Values are reported as mean \(\pm\) standard
deviation across three random seeds.}
\label{tab:higher_order_analysis}
\renewcommand{\arraystretch}{1.16}
\small
\begin{tabularx}{\textwidth}{p{2.7cm} Y C C C C}
\hline
\textbf{Dataset} &
\textbf{Action family} &
\textbf{Pairwise} &
\textbf{Static HG} &
\textbf{Proposed} &
\textbf{Gain} \\
\hline

EPIC-KITCHENS-100/VISOR &
Surface-dependent manipulation &
\(59.1 \pm 0.8\) &
\(61.5 \pm 0.7\) &
\(\mathbf{65.3 \pm 0.6}\) &
\(+6.2 \pm 0.7\) \\

EPIC-KITCHENS-100/VISOR &
Tool-mediated manipulation &
\(56.8 \pm 0.9\) &
\(60.4 \pm 0.8\) &
\(\mathbf{65.8 \pm 0.7}\) &
\(+9.0 \pm 0.8\) \\

EPIC-KITCHENS-100/VISOR &
Contact- or state-change actions &
\(58.7 \pm 0.8\) &
\(60.2 \pm 0.7\) &
\(\mathbf{64.1 \pm 0.6}\) &
\(+5.4 \pm 0.7\) \\
\hline

Assembly101 &
Bimanual manipulation &
\(46.1 \pm 0.9\) &
\(49.7 \pm 0.8\) &
\(\mathbf{55.6 \pm 0.7}\) &
\(+9.5 \pm 0.9\) \\

Assembly101 &
Tool-mediated manipulation &
\(44.9 \pm 1.0\) &
\(48.6 \pm 0.9\) &
\(\mathbf{54.4 \pm 0.8}\) &
\(+9.5 \pm 0.9\) \\
\hline
\end{tabularx}
\end{table}

The largest gains occur for tool-mediated and bimanual actions. These classes depend on the joint participation of several entities, such as the active hand, tool, manipulated object, and supporting surface, or both hands acting on the same object. The corresponding family-level results show higher mean \(F_1\) scores
for the tested hypergraph formulation than for the matched pairwise
formulation.

The proposed method also has higher mean scores than the static
hypergraph across the reported families. This indicates that higher-order structure alone is beneficial, but that reconstructing the manipulation configuration from time-varying entity relations provides additional information. The smaller gains for surface-dependent and contact-state actions suggest that strong pairwise cues, such as hand--object contact or object--surface proximity, already provide partial evidence. Overall, the results support the use of time-varying higher-order relations for manipulation classes defined by coordinated multi-entity interactions.

\subsection{Qualitative Hyperedge-Importance Analysis}
\label{subsec:qualitative_results}

We qualitatively examine whether the hyperedges emphasized by the model correspond to meaningful manipulation evidence. For each video, the importance score defined in Section~\ref{subsec:prediction_explanation} ranks the selected hyperedges across temporal steps. The highest-ranked hyperedges are mapped back to their participating entities and video frames. These scores indicate the relational evidence emphasized by the model, but should not be interpreted as causal explanations.

\begin{figure}[!htbp]
    \centering
    \includegraphics[width=\textwidth]{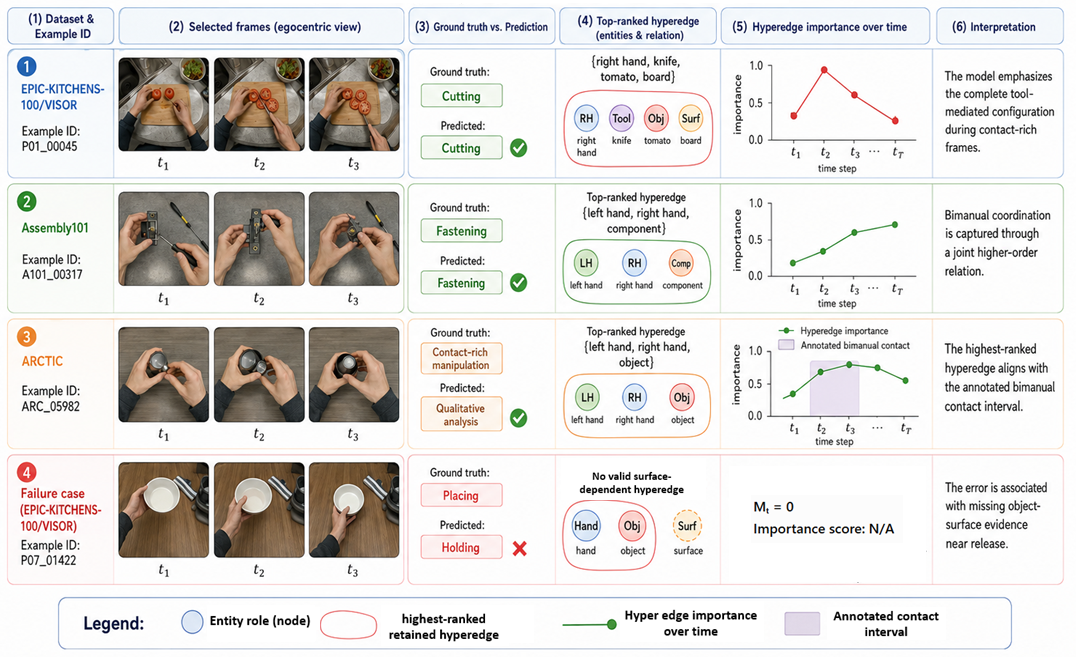}
    \caption{Representative hyperedge-importance visualizations. The
EPIC-KITCHENS-100/VISOR and Assembly101 examples show selected frames,
ground-truth and predicted labels, and the highest-ranked hyperedge.
The ARCTIC example compares normalized hyperedge importance with an
annotated bimanual-contact interval. In the failure case, no valid
hyperedge is instantiated because a required entity is missing.}
    \label{fig:qualitative_explanations}
\end{figure}

Figure~\ref{fig:qualitative_explanations} presents representative examples from EPIC-KITCHENS-100/VISOR, Assembly101, and ARCTIC. In the EPIC-KITCHENS-100/VISOR cutting example, the model assigns the highest importance to the hyperedge connecting the active hand, knife, manipulated object, and cutting surface. Its score increases during tool--object contact, indicating that the model focuses on the complete tool-mediated configuration rather than on the appearance of the knife alone.

For the Assembly101 example, the dominant hyperedge connects the left hand, right hand, and manipulated component. This configuration is informative for distinguishing bimanual actions that contain similar objects and local hand motions but differ in how both hands coordinate around the principal component.

ARCTIC is used only for supporting contact-oriented analysis. The highest-ranked hyperedges are compared with intervals of coordinated bimanual contact provided by the dataset annotations. The observed temporal correspondence indicates that the learned importance scores emphasize interaction periods in which both hands actively participate in manipulating the object. These annotations are used only for analysis and are not supplied to the recognition model.

The failure case illustrates the dependence of the construction on
entity localization. When the supporting surface is not detected, the surface-dependent
templates are not instantiated. If no other manipulation template is
valid at that temporal step, then \(M_t=0\); the prediction consequently
relies on the pooled node representation, and no hyperedge-importance
score is returned for that temporal step.

Overall, the qualitative results show that the model associates its decisions with explicit multi-entity manipulation configurations. Correct predictions are supported by hyperedges that agree with the functional structure of the action, whereas failure cases reveal limitations in entity detection and relation estimation.

\section{Conclusion and Future Directions}
\label{sec:conclusion}

This paper presented a dynamic manipulation hypergraph framework for vision-based human activity recognition. Unlike conventional pairwise graphs, the proposed representation models complete manipulation configurations involving hands, objects, tools, and supporting surfaces as higher-order hyperedges. By combining entity-level visual features, contact and motion predicates, time-varying hypergraph construction, and hypergraph reasoning, the framework is designed to capture relational patterns that are difficult to represent through isolated pairwise connections.

The evaluation considers daily egocentric manipulation and bimanual procedural activities using EPIC-KITCHENS-100/VISOR and Assembly101, while ARCTIC is used for supporting contact-oriented qualitative analysis. Comparisons with video-only, entity-based, matched pairwise-graph, and static-hypergraph baselines examine the respective contributions of explicit relational modeling, higher-order representation, and time-varying hypergraph construction. The family-wise and qualitative analyses further indicate that the proposed representation is particularly suitable for tool-mediated, surface-dependent, bimanual, and contact- or state-change actions.

Several limitations should be acknowledged. First, the constructed hypergraphs depend on reliable localization of hands, objects, tools, and supporting surfaces; missed or inaccurate detections may produce incomplete manipulation configurations. Second, the predefined hyperedge templates cover common manipulation patterns but may not represent all interactions encountered in unconstrained environments. Third, differences in dataset annotations require dataset-specific adaptation of the available entity roles and candidate templates. Finally, the hyperedge importance scores provide attention-based relational evidence rather than causal explanations of the model predictions.

Future work will investigate data-driven hyperedge discovery to reduce dependence on predefined templates. Incorporating open-vocabulary visual and relational representations may further improve generalization to unseen objects, tools, and actions. Other promising directions include robustness to missing entities, cross-dataset transfer, online activity recognition and anticipation, richer object-state modeling, and more rigorous evaluation of explanation faithfulness. These extensions may broaden the applicability of dynamic manipulation hypergraphs while preserving their explicit and interpretable relational structure.

\section*{Declarations}

\textbf{Funding:}
This research received no specific grant from any funding agency in the
public, commercial, or not-for-profit sectors.

\textbf{Competing interests:}
The author declares that there are no competing interests.

\textbf{Author contributions:}
The author was solely responsible for all aspects of the study, including
conceptualization, methodology development, software implementation,
experimental design, validation, formal analysis, investigation, data
curation, visualization, interpretation of the results, and preparation,
review, and editing of the manuscript.

\textbf{Data availability:}
The datasets used in this study are publicly available from the official
repositories of EPIC-KITCHENS-100, VISOR, Assembly101, and ARCTIC.

\FloatBarrier
\bibliography{sn-bibliography}

@inproceedings{Wang_2023_CVPR_VideoMAEv2,
  author    = {Wang, Limin and Huang, Bingkun and Zhao, Zhiyu and Tong, Zhan and He, Yinan and Wang, Yi and Wang, Yali and Qiao, Yu},
  title     = {VideoMAE V2: Scaling Video Masked Autoencoders With Dual Masking},
  booktitle = {Proceedings of the IEEE/CVF Conference on Computer Vision and Pattern Recognition},
  pages     = {14549--14560},
  year      = {2023}
}

@inproceedings{Wang_2024_ECCV_InternVideo2,
  author    = {Wang, Yi and Li, Kunchang and Li, Xinhao and Yu, Jiashuo and He, Yinan and Wang, Chenting and Chen, Guo and Pei, Baoqi and Zheng, Rongkun and Wang, Zun and Shi, Yansong and Jiang, Tianxiang and Li, Tianxiang and Xu, Jilan and Zhang, Hongjie and Huang, Yifei and Qiao, Yu and Wang, Yali and Wang, Limin},
  title     = {InternVideo2: Scaling Foundation Models for Multimodal Video Understanding},
  booktitle = {Computer Vision -- ECCV 2024},
  pages     = {396--416},
  year      = {2024}
}

@article{Damen_2022_IJCV_EPIC100,
  author  = {Damen, Dima and Doughty, Hazel and Farinella, Giovanni Maria and Furnari, Antonino and Kazakos, Evangelos and Moltisanti, Davide and Munro, Jonathan and Perrett, Toby and Price, Will and Wray, Michael},
  title   = {Rescaling Egocentric Vision: Collection, Pipeline and Challenges for {EPIC-KITCHENS-100}},
  journal = {International Journal of Computer Vision},
  volume  = {130},
  number  = {1},
  pages   = {33--55},
  year    = {2022}
}

@inproceedings{Darkhalil_2022_NeurIPS_VISOR,
  author    = {Darkhalil, Ahmad and Shan, Dandan and Zhu, Bin and Ma, Jian and Kar, Amlan and Higgins, Richard and Fidler, Sanja and Fouhey, David and Damen, Dima},
  title     = {{EPIC-KITCHENS VISOR Benchmark: VIdeo Segmentations and Object Relations}},
  booktitle = {Advances in Neural Information Processing Systems},
  volume    = {35},
  note      = {Datasets and Benchmarks Track},
  year      = {2022}
}

@inproceedings{Perrett_2025_CVPR_HDEPIC,
  author    = {Perrett, Toby and Darkhalil, Ahmad and Sinha, Saptarshi and Emara, Omar and Pollard, Sam and Parida, Kranti Kumar and Liu, Kaiting and Gatti, Prajwal and Bansal, Siddhant and Flanagan, Kevin and Chalk, Jacob and Zhu, Zhifan and Guerrier, Rhodri and Abdelazim, Fahd and Zhu, Bin and Moltisanti, Davide and Wray, Michael and Doughty, Hazel and Damen, Dima},
  title     = {HD-EPIC: A Highly-Detailed Egocentric Video Dataset},
  booktitle = {Proceedings of the IEEE/CVF Conference on Computer Vision and Pattern Recognition},
  pages     = {23901--23913},
  year      = {2025}
}

@inproceedings{Yan_2018_AAAI_STGCN,
  author    = {Yan, Sijie and Xiong, Yuanjun and Lin, Dahua},
  title     = {Spatial Temporal Graph Convolutional Networks for Skeleton-Based Action Recognition},
  booktitle = {Proceedings of the AAAI Conference on Artificial Intelligence},
  year      = {2018}
}

@inproceedings{Wang_2023_CVPR_3Mformer,
  author    = {Wang, Lei and Koniusz, Piotr},
  title     = {3Mformer: Multi-Order Multi-Mode Transformer for Skeletal Action Recognition},
  booktitle = {Proceedings of the IEEE/CVF Conference on Computer Vision and Pattern Recognition},
  pages     = {5620--5631},
  year      = {2023}
}

@inproceedings{Zhou_2024_CVPR_BlockGCN,
  author    = {Zhou, Yuxuan and Yan, Xudong and Cheng, Zhi-Qi and Yan, Yan and Dai, Qi and Hua, Xian-Sheng},
  title     = {BlockGCN: Redefine Topology Awareness for Skeleton-Based Action Recognition},
  booktitle = {Proceedings of the IEEE/CVF Conference on Computer Vision and Pattern Recognition},
  pages     = {2049--2058},
  year      = {2024}
}

@article{Zhou_2022_arXiv_Hyperformer,
  author  = {Zhou, Yuxuan and Cheng, Zhi-Qi and Li, Chao and Fang, Yanwen and Geng, Yifeng and Xie, Xuansong and Keuper, Margret},
  title   = {Hypergraph Transformer for Skeleton-Based Action Recognition},
  journal = {arXiv preprint arXiv:2211.09590},
  year    = {2022}
}

@inproceedings{Ray_2025_WACV_AutoregAdHGFormer,
  author    = {Ray, Abhisek and Raj, Ayush and Kolekar, Maheshkumar H.},
  title     = {Autoregressive Adaptive Hypergraph Transformer for Skeleton-Based Activity Recognition},
  booktitle = {Proceedings of the IEEE/CVF Winter Conference on Applications of Computer Vision},
  pages     = {9690--9699},
  year      = {2025}
}

@inproceedings{Nguyen_2025_CVPR_HyperGLM,
  author    = {Nguyen, Trong-Thuan and Nguyen, Pha and Cothren, Jackson and Yilmaz, Alper and Luu, Khoa},
  title     = {HyperGLM: HyperGraph for Video Scene Graph Generation and Anticipation},
  booktitle = {Proceedings of the IEEE/CVF Conference on Computer Vision and Pattern Recognition},
  pages     = {29150--29160},
  year      = {2025}
}

@inproceedings{Kwon_2021_ICCV_H2O,
  author    = {Kwon, Taein and Tekin, Bugra and St{\"u}hmer, Jan and Bogo, Federica and Pollefeys, Marc},
  title     = {H2O: Two Hands Manipulating Objects for First Person Interaction Recognition},
  booktitle = {Proceedings of the IEEE/CVF International Conference on Computer Vision},
  year      = {2021}
}

@inproceedings{Sener_2022_CVPR_Assembly101,
  author    = {Sener, Fadime and Chatterjee, Dibyadip and Shelepov, Daniel and He, Kun and Singhania, Dipika and Wang, Robert and Yao, Angela},
  title     = {Assembly101: A Large-Scale Multi-View Video Dataset for Understanding Procedural Activities},
  booktitle = {Proceedings of the IEEE/CVF Conference on Computer Vision and Pattern Recognition},
  pages     = {21096--21106},
  year      = {2022}
}

@inproceedings{Liu_2022_CVPR_HOI4D,
  author    = {Liu, Yunze and Liu, Yun and Jiang, Che and Lyu, Kangbo and Wan, Weikang and Shen, Hao and Liang, Boqiang and Fu, Zhoujie and Wang, He and Yi, Li},
  title     = {HOI4D: A 4D Egocentric Dataset for Category-Level Human-Object Interaction},
  booktitle = {Proceedings of the IEEE/CVF Conference on Computer Vision and Pattern Recognition},
  pages     = {21013--21022},
  year      = {2022}
}

@inproceedings{Fan_2023_CVPR_ARCTIC,
  author    = {Fan, Zicong and Taheri, Omid and Tzionas, Dimitrios and Kocabas, Muhammed and Kaufmann, Manuel and Black, Michael J. and Hilliges, Otmar},
  title     = {ARCTIC: A Dataset for Dexterous Bimanual Hand-Object Manipulation},
  booktitle = {Proceedings of the IEEE/CVF Conference on Computer Vision and Pattern Recognition},
  pages     = {12943--12954},
  year      = {2023}
}

@inproceedings{Shiota_2024_WACV_ContactState,
  author    = {Shiota, Tsukasa and Takagi, Motohiro and Kumagai, Kaori and Seshimo, Hitoshi and Aono, Yushi},
  title     = {Egocentric Action Recognition by Capturing Hand-Object Contact and Object State},
  booktitle = {Proceedings of the IEEE/CVF Winter Conference on Applications of Computer Vision},
  pages     = {6527--6537},
  year      = {2024}
}

@inproceedings{Rodin_2024_CVPR_ActionSceneGraphs,
  author    = {Rodin, Ivan and Furnari, Antonino and Min, Kyle and Tripathi, Subarna and Farinella, Giovanni Maria},
  title     = {Action Scene Graphs for Long-Form Understanding of Egocentric Videos},
  booktitle = {Proceedings of the IEEE/CVF Conference on Computer Vision and Pattern Recognition},
  pages     = {18622--18632},
  year      = {2024}
}

@article{Ziaeetabar_2024_MVA_MultiSentence,
  author  = {Ziaeetabar, Fatemeh and Safabakhsh, Reza and Momtazi, Saeedeh and Tamosiunaite, Minija and W{\"o}rg{\"o}tter, Florentin},
  title   = {Multi Sentence Description of Complex Manipulation Action Videos},
  journal = {Machine Vision and Applications},
  volume  = {35},
  number  = {4},
  pages   = {64},
  year    = {2024}
}

@article{Ziaeetabar_2024_IEEEAccess_HierarchicalBimanual,
  author  = {Ziaeetabar, Fatemeh and Tamosiunaite, Minija and W{\"o}rg{\"o}tter, Florentin},
  title   = {A Hierarchical Graph-Based Approach for Recognition and Description Generation of Bimanual Actions in Videos},
  journal = {IEEE Access},
  volume  = {12},
  pages   = {180328--180360},
  year    = {2024}
}

@article{Ziaeetabar_2025_IEEEAccess_AdaptiveMultimodal,
  author  = {Ziaeetabar, Fatemeh and W{\"o}rg{\"o}tter, Florentin},
  title   = {Adaptive Multimodal Graph Reasoning With Foundation Models for Fine-Grained Action Recognition},
  journal = {IEEE Access},
  volume  = {13},
  pages   = {201990--202009},
  year    = {2025}
}

@inproceedings{Feng_2019_AAAI_HGNN,
  title     = {Hypergraph Neural Networks},
  author    = {Feng, Yifan and You, Haoxuan and Zhang, Zizhao and Ji, Rongrong and Gao, Yue},
  booktitle = {Proceedings of the AAAI Conference on Artificial Intelligence},
  year      = {2019}
}

@article{Dessalene_2021_arXiv_EgoOMG,
  author  = {Dessalene, Eadom and Devaraj, Chinmaya and Maynord, Michael
             and Fermuller, Cornelia and Aloimonos, Yiannis},
  title   = {Forecasting Action Through Contact Representations From First Person Video},
  journal = {IEEE Transactions on Pattern Analysis and Machine Intelligence},
  volume  = {45},
  number  = {6},
  pages   = {6703--6714},
  year    = {2023},
  doi     = {10.1109/TPAMI.2021.3055233}
}

@inproceedings{Zhang_2022_arXiv_EgoHOS,
  author    = {Zhang, Lingzhi and Zhou, Shenghao and Stent, Simon and Shi, Jianbo},
  title     = {Fine-Grained Egocentric Hand-Object Segmentation:
               Dataset, Model, and Applications},
  booktitle = {Computer Vision -- ECCV 2022},
  pages     = {127--145},
  year      = {2022},
  doi       = {10.1007/978-3-031-19818-2_8}
}

@inproceedings{Liu_2024_ECCV_GroundingDINO,
  author    = {Liu, Shilong and Zeng, Zhaoyang and Ren, Tianhe and
               Li, Feng and Zhang, Hao and Yang, Jie and Li, Chunyuan and
               Yang, Jianwei and Su, Hang and Zhu, Jun and Zhang, Lei},
  title     = {Grounding DINO: Marrying DINO with Grounded Pre-Training
               for Open-Set Object Detection},
  booktitle = {Computer Vision -- ECCV 2024},
  year      = {2024}
}

@inproceedings{Ravi_2025_ICLR_SAM2,
  author    = {Ravi, Nikhila and Gabeur, Valentin and Hu, Yuan-Ting and
               Hu, Ronghang and Ryali, Chaitanya and Ma, Tengyu and
               Khedr, Haitham and R{\"a}dle, Roman and Rolland, Chloe and
               Gustafson, Laura and Mintun, Eric and Pan, Junting and
               Alwala, Kalyan Vasudev and Carion, Nicolas and Wu, Chao-Yuan
               and Girshick, Ross and Doll{\'a}r, Piotr and
               Feichtenhofer, Christoph},
  title     = {{SAM 2}: Segment Anything in Images and Videos},
  booktitle = {International Conference on Learning Representations},
  year      = {2025}
}

@article{ziaeetabar2025efficientgformer,
  title         = {{EfficientGFormer}: Multimodal Brain Tumor Segmentation via Pruned Graph-Augmented Transformer},
  author        = {Ziaeetabar, Fatemeh},
  journal       = {arXiv preprint arXiv:2508.01465},
  year          = {2025},
  eprint        = {2508.01465},
  archivePrefix = {arXiv},
  primaryClass  = {cs.CV},
  doi           = {10.48550/arXiv.2508.01465}
}

\end{document}